\documentclass[11pt]{article}
\PassOptionsToPackage{table}{xcolor}

\usepackage[preprint]{acl}

\usepackage{times}
\usepackage{latexsym}

\usepackage[T1]{fontenc}
\usepackage[utf8]{inputenc}
\ifdefined\XeTeXversion
\usepackage{newunicodechar}
\newunicodechar{’}{'}
\fi

\usepackage{microtype}

\usepackage{inconsolata}

\usepackage{graphicx}
\usepackage{amsfonts}
\usepackage{amsmath}

\usepackage{booktabs} %
\usepackage{hyperref}
\usepackage[table]{xcolor}
\usepackage[most,skins]{tcolorbox} %
\usepackage{enumitem}
\tcbuselibrary{skins,breakable}
\usepackage{tabularx}
\usepackage{array}
\definecolor{citebluish}{HTML}{397fbe}
\hypersetup{
  colorlinks=true,
  linkcolor=citebluish,
  citecolor=citebluish,
  urlcolor=citebluish
}

\definecolor{mygreen}{HTML}{298D66}
\definecolor{myred}{HTML}{F94153}
\definecolor{tts_or}{HTML}{8AD879}
\definecolor{tts_sc}{HTML}{53D2DC}
\definecolor{grpo}{HTML}{3C5488}
\definecolor{gspo}{HTML}{8491B4}
\definecolor{srpo}{HTML}{4DBBD5}
\definecolor{octopus}{HTML}{669999}
\newcolumntype{L}{>{\hspace{2.2pt}}l<{\hspace{2.2pt}}}
\newcolumntype{Y}{>{\hspace{2.2pt}\centering\arraybackslash}X<{\hspace{2.2pt}}}

\tcbset{
  promptbox/.style={
    width=\linewidth,
    top=8pt,
    bottom=2pt,
    left=4pt,
    right=4pt,
    colback=gray!1!white,
    colframe=black,
    colbacktitle=black,
    enhanced,
    center,
    fontupper=\normalsize,
    before upper={\setlength{\baselineskip}{1.12\baselineskip}}, %
    attach boxed title to top left={yshift=-0.1in,xshift=0.15in},
    boxed title style={boxrule=0pt,colframe=white,},
  }
}

\newtcolorbox{promptbox}[2][]{promptbox,title=#2,#1}
\newcounter{case}
\newtcolorbox{casebox}[2][]{%
  before upper={\refstepcounter{case}},
  colframe=black!80!black,
  colback=gray!10!white,
  colbacktitle=black!80!white,
  coltitle=white,
  boxrule=0.5mm,
  enhanced,
  rounded corners,
  width=\linewidth,
  title=\textbf{Case Study #2},
  #1
}

\title{Seeing with You: Perception-Reasoning Co-evolution for Multimodal Reasoning}

\author{
\normalfont
Ziqi Miao$^{1,\star}$,
Haonan Jia$^{2,\star}$,
Lijun Li$^{1,\star\dagger}$,
Chen Qian$^3$,
Yuan Xiong$^1$ \\
\normalfont
Wenting Yan$^4$,
Yongyao Wang$^5$,
Zaoying Huang$^5$,
Ken Deng$^5$,
Jing Shao$^{1,\dagger}$ \\
$^1$ Shanghai Artificial Intelligence Laboratory, Shanghai, China \\
$^2$ Beihang University, Beijing, China \\
$^3$ Gaoling School of Artificial Intelligence, Renmin University of China, Beijing, China \\
$^4$ Zhejiang University, Hangzhou, China \\
$^5$ Independent Researcher \\
\texttt{\{lilijun,shaojing\}@pjlab.org.cn}
}

\hypersetup{pdfauthor={Ziqi Miao, Haonan Jia, Lijun Li, Chen Qian, Yuan Xiong, Wenting Yan, Yongyao Wang, Zaoying Huang, Ken Deng, Jing Shao}}

\begin{document}
\maketitle
\begingroup
\renewcommand{\thefootnote}{}
\makeatletter
\renewcommand{\@makefntext}[1]{\noindent #1}
\makeatother
\footnotetext{%
$^{\star}$ Equal contribution.\\
$^{\dagger}$ Corresponding authors.\\
Code: \url{https://github.com/Dtc7w3PQ/PRCO}}
\endgroup
\begin{abstract}
Reinforcement learning with verifiable rewards (RLVR) has substantially enhanced the reasoning capabilities of multimodal large language models (MLLMs). However, existing RLVR approaches typically rely on outcome-driven optimization that updates both perception and reasoning using a shared reward based solely on the final answer. This shared reward blurs credit assignment, frequently improving reasoning patterns while failing to reliably enhance the accuracy of upstream visual evidence extraction. To address this perception bottleneck, we introduce \textbf{PRCO} (\textbf{P}erception-\textbf{R}easoning \textbf{Co}-evolution), a dual-role RLVR framework with a shared policy. PRCO consists of two cooperative roles:
an Observer that generates an evidence caption tailored to the question
and a Solver that predicts the final answer based on this caption.
Crucially, PRCO employs role-specific reward signals: the Solver is optimized using verifiable outcome rewards on the final answer, while the Observer receives a utility reward derived from the Solver's downstream success. Extensive experiments across eight challenging multimodal reasoning benchmarks demonstrate that PRCO yields consistent improvements across model scales by over 7 points on average accuracy compared to the base model, outperforming prior open-source RL-tuned baselines.
\end{abstract}

\section{Introduction}
\label{sec:intro}

Reinforcement learning with verifiable rewards (RLVR), particularly online algorithms such as Group Relative Policy Optimization (GRPO), has substantially enhanced the reasoning capabilities of Large Language Models (LLMs) in text-only problem domains~\citep{deepseek2025r1,kimi2025k1_5,xu2026stable,huang2026step}.
Building on these advances, recent work extends RLVR to Multimodal Large Language Models (MLLMs) for challenging multimodal reasoning tasks~\citep{huang2025visionr1,shen2025vlmr1,wang2025vlrethinker,zhu2025activeo3,zheng2025deepeyes,su2025openthinkimg,fan2025grit,cao2025groundr1,meng2025mmeureka}.
Most prior multimodal RLVR work focuses on data-centric curation~\citep{liu2025noisyrollout,leng2025mmr1,wu2025synthrl,li2025truthfew,wang2025sotaless} and reward-centric engineering~\citep{liu2025visualrft,wan2025srpo,su2025pixelreasoner,wang2025vicrit}.
Effective reasoning relies on accurate perception, which provides the necessary grounding for logical deduction~\citep{liu2025more}. However, RLVR is often applied in an outcome-driven manner, verifying only the final textual answer while largely neglecting the accuracy of upstream visual perception~\citep{wang2025papo,li2025selfrewarding}.

\begin{figure}[t]
    \centering
    \includegraphics[width=\columnwidth]{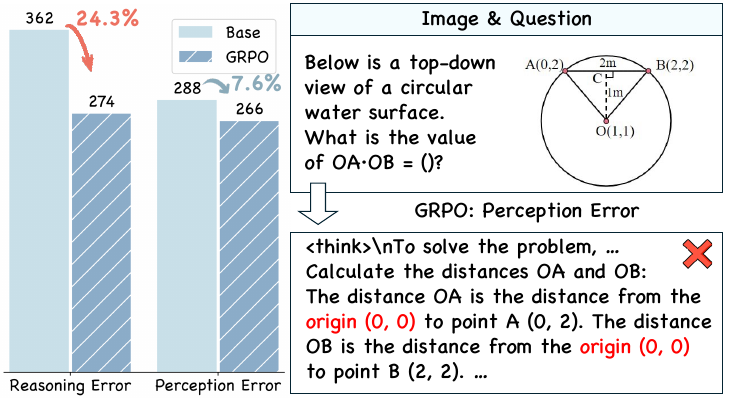}
    \caption{Diagnostic analysis of GRPO on WeMath~\cite{qiao2024wemath}. Left: GRPO reduces reasoning errors much more than perception errors. Right: a representative failure case caused by incorrect perception.}
    \label{fig:grpo_diagnostic}
\end{figure}

To concretely examine this bottleneck, we conduct a diagnostic analysis using GRPO as a representative multimodal RLVR baseline.
We train Qwen2.5-VL-7B~\citep{bai2025qwen25vl} with GRPO and compare its failure modes against its initialization on WeMath~\citep{qiao2024wemath} using fine-grained error categorization.
As shown in Fig.~\ref{fig:grpo_diagnostic}, \textit{training with GRPO substantially reduces reasoning errors, whereas perception errors improve only marginally over the base model.}
We attribute this bottleneck to outcome-only RLVR with a shared reward updating both perception and reasoning. This blurs credit assignment and improves reasoning patterns without reliably improving visual evidence extraction.

Recently, some works have recognized perception as critical and explored perception-centric RLVR for MLLMs.
These efforts focus on three directions: introducing additional perception-oriented optimization objectives~\citep{xiao2025perceptionr1,wang2025papo}, using weighted token-level credit assignment for perception and reasoning tokens~\citep{huang2025vppo}, requiring an explicit image-description step before reasoning~\citep{xing2025caprl,gou2025racro}.
Despite being promising, these approaches either retain coupled perception--reasoning optimization or assess intermediate visual evidence with a fixed external reasoner. PRCO instead provides adaptive, role-specific feedback between a co-evolving Observer and Solver.
These findings naturally prompt the question: 

\textit{Can we use reliable, separate learning signals for perception and reasoning to decouple them at the gradient level?}

To this end, we introduce \textbf{PRCO} (\textbf{P}erception-\textbf{R}easoning \textbf{Co}-evolution), a dual-role RLVR framework that disentangles perception and reasoning. In this setup, a shared multimodal policy alternates between two roles: 
the Observer, which performs question-conditioned evidence extraction and writes an evidence caption tailored to the question;
and the Solver, which produces the answer based on the question and the caption, using the image when available.
Crucially, they are trained with separate and reliable learning signals: the Solver is optimized with verifiable outcome rewards on the final answer, while the Observer is optimized with a utility reward defined by the Solver's verifier-validated success when conditioned on its caption.
These role-specific learning signals decouple policy updates for perception and reasoning while still allowing joint optimization under a shared policy.
As a result, PRCO drives co-evolution, in which the Observer progressively improves question-grounded visual perception by extracting and articulating question-relevant evidence, while the Solver learns to reason more reliably under explicit evidence guidance.

To validate PRCO, we conduct extensive experiments on eight challenging multimodal reasoning benchmarks covering mathematics, geometry, logic, and multidisciplinary reasoning.
PRCO yields consistent gains across model scales: our 7B model improves average accuracy by 7.18 points over the corresponding base model and outperforms prior open-source RL-tuned methods, while our 3B model improves average accuracy by 7.65 points.
Moreover, on WeMath, PRCO reduces perception errors by 39.2\% relative to the base model, whereas vanilla GRPO achieves only a 7.6\% reduction.

To sum up, our main contributions are threefold:

\begin{itemize}
    \item We propose \textbf{PRCO}, a dual-role RLVR framework that disentangles perception and reasoning with an Observer and a Solver under a shared policy.
    
    \item We demonstrate the effectiveness of PRCO on diverse and challenging multimodal reasoning benchmarks, showing consistent improvements over strong RLVR baselines.
    
    \item We provide extensive ablation and diagnostic analyses that validate PRCO’s key design choices and characterize its effects on perception and reasoning.
\end{itemize}

\section{Method}
\label{sec:method}

\begin{figure*}[t]
  \centering
  \includegraphics[width=\linewidth]{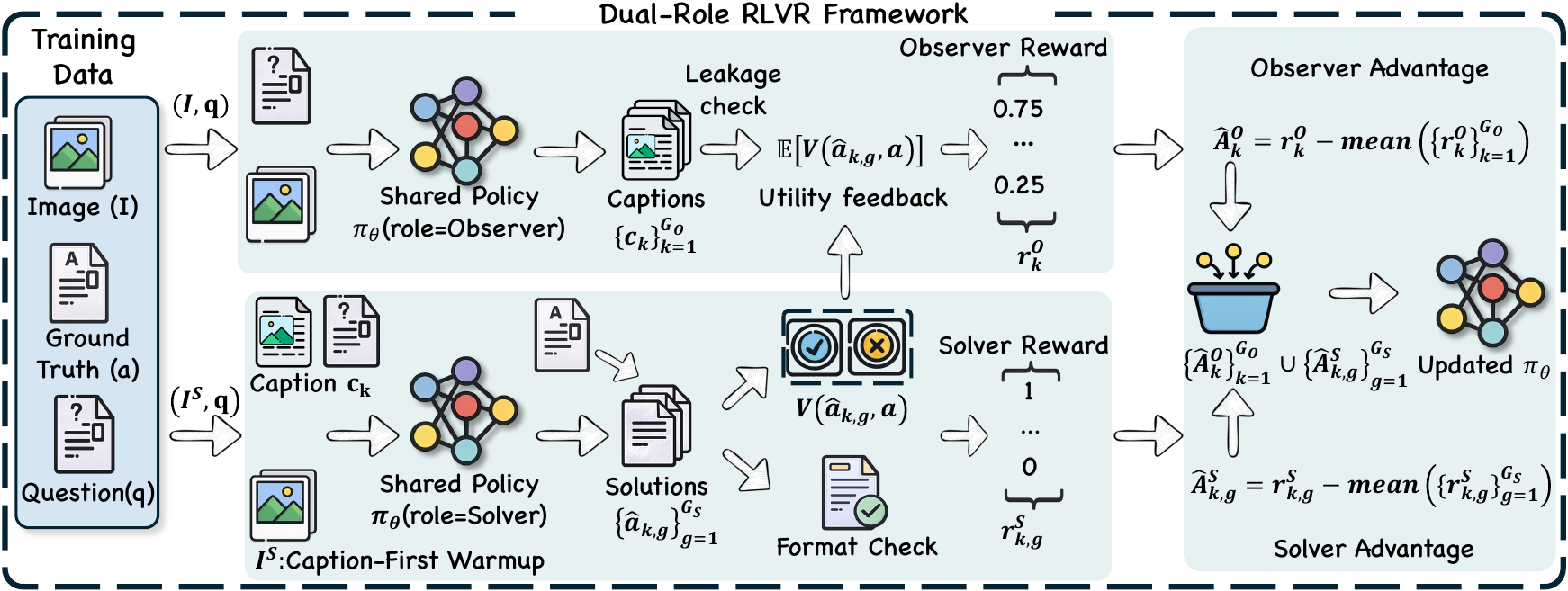}
    \caption{Overview of PRCO. A shared policy alternates between an Observer for question-conditioned evidence captioning and a Solver for evidence-conditioned reasoning. The two roles are jointly optimized with role-specific learning signals and group-relative advantages, enabling perception-reasoning co-evolution under a shared policy.}
  \label{fig:framework}
\end{figure*}

We propose a dual-role RLVR framework that trains a shared MLLM policy in two cooperative roles.
The Observer produces a question-conditioned evidence caption $c$, i.e., one tailored to the question, and the Solver answers based on the caption and the image when available.
We first review GRPO (Sec.~\ref{sec:grpo}), then describe the dual-role interaction loop (Sec.~\ref{sec:overview}), detail the Observer and Solver (Sec.~\ref{sec:observer}--\ref{sec:solver}), and finally present unified optimization with role-specific advantages (Sec.~\ref{sec:unified_opt}).

\subsection{Preliminary: Group Relative Policy Optimization}
\label{sec:grpo}

GRPO~\citep{shao2024deepseekmath} is a reinforcement learning algorithm for fine-tuning a policy LLM without learning a separate value function.
Its key idea is to compute \emph{relative} learning signals by normalizing rewards against other responses sampled from the same prompt.
For a given prompt $x$, the policy generates a group of $G$ complete responses $\{y_i\}_{i=1}^{G}$.
Each response receives a scalar reward $r_i$.
GRPO converts these rewards into response-level advantages via z-score normalization:
\begin{equation}
\hat{A}_i
=
\frac{
r_i - \mathrm{mean}(\{r_j\}_{j=1}^{G})
}{
\mathrm{std}(\{r_j\}_{j=1}^{G}) + \epsilon_{\mathrm{norm}}
},
\label{eq:grpo_adv}
\end{equation}
where $\epsilon_{\mathrm{norm}}$ is a small constant for numerical stability.

\paragraph{Policy update.}
GRPO updates the policy using a PPO-style clipped surrogate objective to improve stability.
To prevent excessive policy drift, the objective is regularized with a KL-divergence penalty to the old policy:
\begin{equation}
\begin{aligned}
\mathcal{L}_{\mathrm{GRPO}}(\theta)
=&-\mathbb{E}_{i,t}\Big[
\min\Big(
\rho_{i,t}(\theta)\hat{A}_i,\;
\\
&\qquad
\mathrm{clip}\big(\rho_{i,t}(\theta),1-\epsilon,1+\epsilon\big)\,\hat{A}_i
\Big)
\Big]
\\
&+\beta\,\mathbb{E}_{x}\Big[
\mathrm{KL}\big(\pi_\theta(\cdot|x)\,\|\,
\pi_{\theta_{\mathrm{old}}}(\cdot|x)\big)
\Big].
\end{aligned}
\label{eq:grpo_loss}
\end{equation}
where $\rho_{i,t}(\theta)$ is the token-level importance ratio, $\epsilon$ is the clipping threshold, and $\beta$ controls the KL penalty.
Optimizing this objective increases the likelihood of responses with positive relative advantages, while the KL term constrains divergence from $\pi_{\theta_{\mathrm{old}}}$ for stable training.

\subsection{Overview}
\label{sec:overview}

\paragraph{RLVR setting.}
Following typical RLVR setups \citep{wang2025vlrethinker,yu2025dapo}, our training dataset $\mathcal{D}$ consists of tuples $(I,q,a)$ where $I$ is an image, $q$ is a question, and $a$ is a short ground-truth answer.
We do not rely on any existing chain-of-thought data and initiate RL training directly without supervised fine-tuning.
We use a lightweight rule-based verifier $V(\hat{a},a)\in\{0,1\}$ that checks whether the predicted answer $\hat{a}$ matches $a$, and a simple format checker $\mathrm{FormatScore}(\hat{a})\in[0,1]$ that evaluates whether the output satisfies the required format.

\paragraph{Two roles under one shared policy.}
We instantiate a single policy $\pi_\theta$ in two roles via role-specific prompting.
We denote by $r_O$ and $r_S$ the prompts for the Observer and Solver, respectively.
For each sample $(I,q,a)$, the Observer first generates an intermediate caption $c$ summarizing question-relevant visual evidence; then the Solver outputs the final answer conditioned on $c$ and optionally the image.
We denote the Solver's visual input as $I^{S}\in\{\emptyset, I\}$.
For each training instance, we sample $G_O$ candidate captions $\{c_k\}_{k=1}^{G_O}$ under the Observer role; given a caption $c_k$, we sample $G_S$ candidate answers $\{\hat{a}_{k,g}\}_{g=1}^{G_S}$ under the Solver role.
The Observer is encouraged to externalize visual evidence into captions, while the Solver is trained to leverage captions for evidence-conditioned reasoning.

\subsection{Observer: Utility-Driven Evidence Captioning}
\label{sec:observer}

The Observer converts high-dimensional visual input into a textual evidence signal by producing a question-conditioned evidence caption that summarizes the visual evidence most relevant to $q$ (e.g., entities, attributes, and relations). Formally, given $(I,q)$, the Observer samples $c \sim \pi_\theta(\cdot \mid I,q,r_O)$.

\paragraph{Utility reward with leakage suppression.}
Directly verifying the intrinsic quality of an evidence caption is difficult. We therefore train the Observer through the downstream utility that its caption provides to the Solver. A key failure mode is answer shortcutting, where the Observer directly places the final answer in the caption instead of extracting question-relevant visual evidence.
To suppress this behavior, we use an auxiliary LLM-based leakage checker that takes the caption $c$ and question $q$ as input and outputs a binary leakage indicator.
Let $\mathbb{I}_{\mathrm{leak}}(q,c)\in\{0,1\}$ be the indicator of answer leakage in $c$, where 1 indicates the presence of leakage.
For a sampled caption $c_k$, we define the Observer reward as
\begin{equation}
r_k^{O}
=
\bigl(1-\mathbb{I}_{\mathrm{leak}}(q,c_k)\bigr)\,
\mathbb{E}_{\hat{a}\sim \pi_\theta(\cdot \mid I^{S}, q, c_k, r_S)}
\left[
V(\hat{a}, a)
\right].
\label{eq:observer_reward}
\end{equation}
The expectation is approximated by the empirical mean of the verifier scores over $G_S$ sampled Solver rollouts conditioned on $(I^{S}, q, c_k, r_S)$. This reward favors captions that help downstream solving and suppresses captions judged as leaking the answer.

\subsection{Solver: Evidence-Conditioned Reasoning}
\label{sec:solver}

The Solver produces a short final answer by reasoning over the question and the Observer's caption, with the image provided when available.
Conditioning on $c$ encourages explicit evidence-driven reasoning, while image input helps recover global structure or complex geometric relations that are difficult to fully convey in text.
Formally, given a caption $c$, the Solver samples $\hat{a} \sim \pi_\theta(\cdot \mid I^{S},q,c,r_S)$. 

\paragraph{Solver reward.}
We define the correctness reward via the verifier as $r^{\mathrm{acc}} = V(\hat{a},a)$.
In addition, $r^{\mathrm{format}}\in[0,1]$ measures whether the response strictly follows the required format. We compute it with a simple rule-based checker as $r^{\mathrm{format}}=\mathrm{FormatScore}(\hat{a})$.
The Solver is rewarded for both correctness and basic format compliance:
\begin{equation}
r^{S}
=
\lambda\, r^{\mathrm{acc}}
+
(1-\lambda)\, r^{\mathrm{format}},
\label{eq:solver_reward}
\end{equation}
where $\lambda$ balances accuracy and format compliance, with a default value of $0.9$.

\paragraph{Caption-first warmup.}
In early training, we set $I^{S}=\emptyset$ so that the Solver must rely on $(q,c)$.
We find that if the Solver receives both the image and the caption too early, it tends to ignore the caption and solve directly from the image, which can drown out the learning signal for improving captions.
Therefore, we first warm up the Solver to solve using only captions; after a short warmup, we switch to $I^{S}=I$ to restore full multimodal grounding while retaining the benefits of caption conditioning.

\subsection{Unified Policy Optimization with Role-Specific Advantages}
\label{sec:unified_opt}

The Observer and Solver share the same policy and are optimized jointly. However, their rollouts define different comparison groups for relative optimization. Observer captions are compared across samples from the same $(I,q)$ instance, whereas Solver answers are compared within caption-conditioned answer groups.
We therefore compute relative advantages separately for the two roles and optimize the shared policy over the combined rollouts.

\paragraph{Role-wise grouping and advantage computation.}
For each sample $(I,q,a)$, the Observer generates $G_O$ candidate captions $\{c_k\}_{k=1}^{G_O}$ under $\pi_\theta(\cdot \mid I,q,r_O)$.
For each caption $c_k$, we generate $G_S$ Solver answers $\{\hat{a}_{k,g}\}_{g=1}^{G_S}$ under $(I^{S},q,c_k,r_S)$ and compute rewards $\{r^{S}_{k,g}\}_{g=1}^{G_S}$ using Eq.~\eqref{eq:solver_reward}.
We then compute Observer rewards $\{r^{O}_k\}_{k=1}^{G_O}$ via Eq.~\eqref{eq:observer_reward}.
Following Eq.~\eqref{eq:grpo_adv}, we compute group-relative advantages separately for the two roles while omitting the standard deviation normalization term. Concretely, for a reward group $\{r_i\}_{i=1}^{G}$, we use $\hat{A}_i = r_i - \mathrm{mean}(\{r_j\}_{j=1}^{G})$.
Centering removes each role-specific group's reward offset while preserving its observed reward differences without rescaling by the group standard deviation.
We compute caption advantages $\{\hat{A}^{O}_k\}_{k=1}^{G_O}$ from $\{r^{O}_k\}_{k=1}^{G_O}$ across the $G_O$ captions.
For the Solver update, we reuse these evidence-conditioned answer rollouts.
For each $(I,q,a)$, we preferentially sample one caption index $\tilde{k}$ uniformly from captions with $\mathrm{Var}(\{r^{S}_{k,g}\}_{g=1}^{G_S})>0$ to avoid degenerate relative signals, and compute Solver advantages $\{\hat{A}^{S}_{\tilde{k},g}\}_{g=1}^{G_S}$ from $\{r^{S}_{\tilde{k},g}\}_{g=1}^{G_S}$.

\paragraph{Unified policy update.}
We aggregate Observer caption trajectories associated with $\hat{A}^{O}$ and Solver answer trajectories associated with $\hat{A}^{S}$ into a combined rollout batch and optimize a unified GRPO-style objective:
\begin{equation}
\mathcal{L}_{\mathrm{dual}}(\theta)
=
\mathcal{L}_{\mathrm{GRPO}}\big(\theta;\hat{A}^{S}\big)
+
\mathcal{L}_{\mathrm{GRPO}}\big(\theta;\hat{A}^{O}\big),
\label{eq:dual_role_grpo}
\end{equation}
where $\mathcal{L}_{\mathrm{GRPO}}(\theta;\hat{A})$ denotes Eq.~\eqref{eq:grpo_loss} instantiated on the corresponding role trajectories with advantage $\hat{A}$. Following DAPO~\citep{yu2025dapo}, we set $\beta=0$ to remove the KL penalty and encourage exploration.
Under the shared policy, this unified objective jointly trains perception and reasoning.

\section{Experiments}

\begin{table*}[t]
    \centering
    \fontsize{8pt}{9pt}\selectfont
    \setlength{\tabcolsep}{0pt}          %
    \renewcommand{\arraystretch}{1.08}
    \begin{tabularx}{\textwidth}{@{} L *{9}{Y} @{}}
        \toprule
        \textbf{Model} &
        \multicolumn{5}{c}{\textbf{Math-Related}} &
        \multicolumn{3}{c}{\textbf{General Task}} &
        \textbf{Overall} \\
        \cmidrule(lr){2-6}\cmidrule(lr){7-9}\cmidrule(lr){10-10}
        & MathVerse & MathVision & MathVista & WeMath & DynaMath & LogicVista & MMMU-Pro & MMStar & Avg. \\
        \midrule

        \rowcolor{gspo!10}
        \multicolumn{10}{c}{\textbf{\emph{Based on Qwen2.5-VL-3B}}} \\
        \midrule
        Base Model      & 34.13 & 22.50 & 65.00 & 23.52 & 12.37 & 38.70 & 26.76 & 56.06 & 34.88 \\
        GRPO            & 36.29 & 24.70 & 67.40 & 30.57 & 17.96 & 38.92 & 29.88 & 58.00 & 37.97 \\
        DAPO            & \underline{40.98} & \textbf{27.40} & \underline{70.20} & 35.14 & \underline{20.35} & 43.62 & 31.73 & 60.40 & \underline{41.23} \\
        \midrule
        PAPO-G-3B       & 37.56 & 23.61 & 67.60 & 31.05 & 17.96 & 40.26 & 29.01 & 58.86 & 38.24 \\
        PAPO-D-3B       & 40.48 & 26.61 & 69.30 & 32.29 & 19.36 & \textbf{47.20} & 31.27 & \underline{60.66} & 40.90 \\
        MMR1-3B-RL         & 38.57 & 22.17 & 64.50 & \underline{38.29} & 16.36 & 40.49 & 30.34 & 57.20 & 38.49 \\
        Vision-SR1-3B   & 36.29 & 25.65 & 64.90 & 34.48 & 18.16 & 41.16 & \textbf{33.12} & 56.73 & 38.81 \\
        \rowcolor{mygreen!10}\textbf{PRCO-3B}
                        & \textbf{42.51} & \underline{27.27} & \textbf{70.30} & \textbf{40.00} & \textbf{22.36} & \underline{44.97} & \underline{31.85} & \textbf{61.00} & \textbf{42.53} \\
        \midrule

        \rowcolor{gspo!10}
        \multicolumn{10}{c}{\textbf{\emph{Based on Qwen2.5-VL-7B}}} \\
        \midrule
        Base Model          & 43.02 & 25.46 & 70.20 & 35.43 & 20.35 & 45.41 & 35.49 & 64.26 & 42.45 \\
        GRPO                & 44.28 & 28.28 & 75.40 & 41.43 & 25.14 & 46.08 & 39.01 & 64.06 & 45.46 \\
        DAPO                & \underline{48.73} & 29.30 & 74.80 & 45.62 & 26.14 & 47.87 & 41.38 & 65.40 & 47.41 \\
        \midrule
        PAPO-G-7B           & 44.79 & 27.20 & 74.30 & 39.62 & 23.55 & 43.17 & 40.11 & 64.93 & 44.71 \\
        PAPO-D-7B           & 47.33 & 24.34 & \underline{76.70} & 39.05 & 25.34 & 48.54 & \underline{41.50} & 66.93 & 46.22 \\
        R1-ShareVL-7B       & 48.22 & 29.14 & 73.30 & 45.14 & 24.55 & 48.76 & 38.32 & 65.06 & 46.56 \\
        Perception-R1-7B    & 46.70 & 26.74 & 73.40 & 46.48 & 23.75 & 44.07 & 38.20 & 64.33 & 45.46 \\
        NoisyRollout-7B     & 46.44 & 27.50 & 72.30 & 46.10 & 23.15 & 48.32 & 36.30 & 63.93 & 45.51 \\
        MMR1-7B-RL             & 43.90 & 26.01 & 71.60 & \underline{47.87} & 27.14 & \underline{49.44} & 35.08 & 64.80 & 45.73 \\
        VPPO-7B             & 47.20 & \underline{30.52} & 76.60 & 43.81 & \underline{27.94} & 47.87 & 39.65 & \underline{67.20} & \underline{47.60} \\
        Vision-Matters-7B   & 47.08 & 27.23 & 72.30 & 41.71 & 24.75 & 48.09 & 37.10 & 62.20 & 45.06 \\
        Vision-SR1-7B       & 42.76 & 27.76 & 72.30 & 37.14 & 24.55 & 48.32 & 41.38 & 65.26 & 44.93 \\
        \rowcolor{mygreen!10}\textbf{PRCO-7B} & \textbf{49.49} & \textbf{30.86} & \textbf{77.10} & \textbf{50.29} & \textbf{29.74} & \textbf{49.66} & \textbf{42.08} & \textbf{67.80} & \textbf{49.63} \\
        \bottomrule
    \end{tabularx}
    \caption{Main results on eight multimodal reasoning benchmarks with Qwen2.5-VL-3B and Qwen2.5-VL-7B backbones. We report benchmark scores on math-related benchmarks, general-task benchmarks, and their overall average. The best and second-best results within each backbone are highlighted in \textbf{bold} and \underline{underlined}, respectively.}
    \label{tab:results_combined}
\end{table*}

\subsection{Experimental Setup}
\label{sec:exp_setup}

\paragraph{Models, Data, and Baselines.}
We perform direct RL training on the Qwen2.5-VL-3B, Qwen2.5-VL-7B, and Qwen3-VL-8B-Instruct backbones using ViRL39K~\citep{wang2025vlrethinker}.
ViRL39K contains 39K verifiable multimodal reasoning questions across diverse visual formats, such as diagrams and charts.
We benchmark our method against recent open-source reasoning MLLMs at the 3B and 7B scales, and further evaluate it on the stronger Qwen3-VL-8B-Instruct backbone.
For the 3B setting, we compare with PAPO-G-3B and PAPO-D-3B~\citep{wang2025papo}, MMR1-3B-RL~\citep{leng2025mmr1}, and Vision-SR1-3B~\citep{li2025selfrewarding}.
For the 7B setting, we include PAPO-G-7B and PAPO-D-7B~\citep{wang2025papo}, R1-ShareVL-7B~\citep{yao2025r1sharevl}, Perception-R1-7B~\citep{xiao2025perceptionr1}, Vision-Matters-7B~\citep{li2025visionmatters}, NoisyRollout-7B~\citep{liu2025noisyrollout}, MMR1-7B-RL~\citep{leng2025mmr1}, VPPO-7B~\citep{huang2025vppo}, and Vision-SR1-7B~\citep{li2025selfrewarding}.
We also implement two strong RLVR baselines by fine-tuning the Qwen2.5-VL backbones and Qwen3-VL-8B-Instruct with GRPO~\citep{shao2024deepseekmath} and DAPO~\citep{yu2025dapo}. Appendix~\ref{app:qwen3} reports the Qwen3-VL-8B-Instruct results, and Appendix~\ref{app:eval} provides additional details.

\paragraph{Training Details.}
All experiments are implemented using the EasyR1 codebase~\citep{zheng2025easyr1} and optimized with AdamW~\citep{loshchilov2019decoupled}, with a learning rate of $1\times10^{-6}$.
Following PAPO and VPPO~\citep{wang2025papo,huang2025vppo}, we use a rollout batch size of 384. We train all models for 200 optimization steps.
We set the Observer rollout group size to 4. For the Solver, we use a rollout group size of 8, in line with recent multimodal RL training practice~\citep{huang2025vppo,li2025selfrewarding}.
We adopt a caption-first warmup for the first 40 steps, during which the Solver is trained without image inputs to encourage effective caption conditioning before restoring full multimodal inputs.
More training details are provided in Appendix~\ref{app:train}.

\paragraph{Evaluation Benchmarks.}
We evaluate on eight multimodal reasoning benchmarks, including math-related visual reasoning on MathVista~\citep{lu2023mathvista}, MathVerse~\citep{zhang2024mathverse}, MathVision~\citep{wang2024mathvision}, WeMath~\citep{qiao2024wemath}, and DynaMath~\citep{zou2024dynamath}, and general tasks on LogicVista~\citep{xiao2024logicvista}, MMMU-Pro~\citep{yue2024mmmupro}, and MMStar~\citep{chen2024mmstar}.
We use VLMEvalKit~\citep{duan2024vlmevalkit} with greedy decoding for all benchmarks, setting temperature to $0$ and top-p to $1.0$.
We report single-sample greedy results under each benchmark's official VLMEvalKit metric, denoted as accuracy for simplicity.
All models are evaluated under a single fixed evaluation configuration to ensure fair comparison and reproducibility. See Appendix~\ref{app:eval} for additional evaluation details.

\subsection{Main Results}
\textbf{PRCO yields consistent improvements across model scales and task categories.}
As shown in Table~\ref{tab:results_combined}, PRCO improves upon the Qwen2.5-VL backbones at both scales, with average gains of 7.65 and 7.18 points in the 3B and 7B settings, respectively.
Under identical training settings, PRCO outperforms GRPO and DAPO.
PRCO-3B surpasses strong RLVR baselines and recent open-source reasoning MLLMs built on the same 3B backbone.
PRCO-7B further achieves the best overall performance and the strongest results across all evaluated benchmarks, outperforming the strongest baseline, VPPO.
Across task categories, PRCO yields steady gains on math-related benchmarks while also improving general multimodal reasoning, indicating broad cross-task generalization.
Overall, PRCO enables perception-reasoning co-evolution through role-specific, reliable learning signals under a shared policy, setting a new performance bar among open-source MLLMs.

\subsection{Ablation Study}

To better understand the contribution of each component in PRCO, we conduct comprehensive ablations. We report math-related, general-task, and overall averages in Table~\ref{prco_ablation}. More ablation details are provided in Appendix~\ref{app:more}.

\begin{table}[t]
    \centering
    \fontsize{7.5pt}{8.5pt}\selectfont
    \renewcommand{\arraystretch}{1.08}
    \setlength{\tabcolsep}{0.9pt}

    \newcommand{\modeltiny}[1]{{\fontsize{7pt}{8pt}\selectfont #1}}

    \newcolumntype{Y}{>{\centering\arraybackslash}X}

    \begin{tabularx}{\columnwidth}{
        >{\raggedright\arraybackslash}p{0.36\columnwidth}  %
        *{6}{Y}                                            %
    }
        \toprule
        \textbf{Model} &
        \textbf{Math} & $\Delta$ &
        \textbf{General} & $\Delta$ &
        \textbf{Avg} & $\Delta$ \\
        \midrule

        \rowcolor{gspo!10}
        \textbf{Qwen2.5-VL-3B} & 31.50 & - & 40.51 & - & 34.88 & - \\
        + PRCO w/o warmup      & 39.45 & +7.95 & 45.88 & +5.37 & 41.86 & +6.98 \\
        \modeltiny{+ PRCO\,w/o~Observer} & 39.48 & +7.98 & 45.02 & +4.51 & 41.56 & +6.68 \\
        \modeltiny{+ PRCO\,w/o~Solver}   & 33.46 & +1.96 & 41.55 & +1.04 & 36.49 & +1.61 \\
        \rowcolor{mygreen!10}
        \modeltiny{\textbf{+ PRCO}}        & \textbf{40.49} & \textbf{+8.99} & \textbf{45.94} & \textbf{+5.43} & \textbf{42.53} & \textbf{+7.65} \\
        \midrule

        \rowcolor{gspo!10}
        \textbf{Qwen2.5-VL-7B} & 38.89 & - & 48.39 & - & 42.45 & - \\
        + PRCO w/o warmup      & 44.52 & +5.63 & 51.57 & +3.18 & 47.16 & +4.71 \\
        \modeltiny{+ PRCO\,w/o~Observer} & 46.22 & +7.33 & 51.47 & +3.08 & 48.19 & +5.74 \\
        \modeltiny{+ PRCO\,w/o~Solver}   & 41.73 & +2.84 & 49.16 & +0.77 & 44.52 & +2.07 \\
        \rowcolor{mygreen!10}
        \modeltiny{\textbf{+ PRCO}}        & \textbf{47.50} & \textbf{+8.61} & \textbf{53.18} & \textbf{+4.79} & \textbf{49.63} & \textbf{+7.18} \\
        \bottomrule
    \end{tabularx}
    \caption{Ablation study of PRCO on Qwen2.5-VL-3B and Qwen2.5-VL-7B. We report average scores on math-related benchmarks, general-task benchmarks, and all benchmarks; $\Delta$ denotes improvement over the corresponding base model.}
    \label{prco_ablation}
\end{table}

\begin{figure}[t]
    \centering
    \includegraphics[width=\columnwidth]{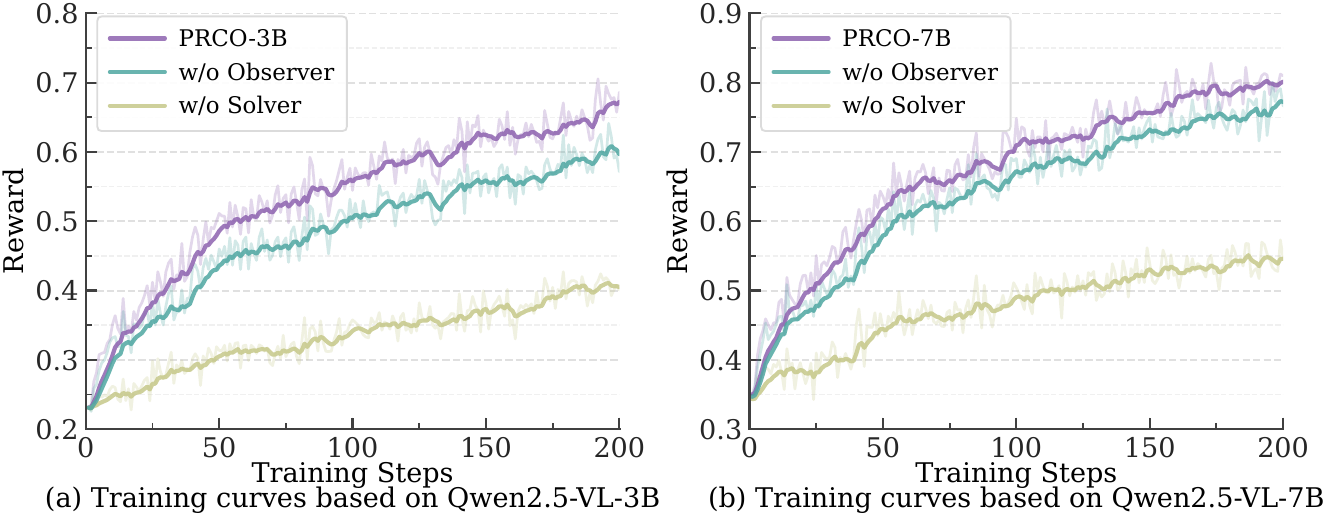}
    \caption{Training reward curves of PRCO and its role-ablation variants with Qwen2.5-VL-3B and Qwen2.5-VL-7B as backbones.}
    \label{fig:ablation_training_curves}
\end{figure}

\paragraph{Effect of role-wise updates.}
We isolate PRCO's role-specific learning signals by dropping one role's trajectories during policy updates while keeping the trajectory generation procedure unchanged.
PRCO w/o Solver updates only from the Observer caption trajectories, whereas PRCO w/o Observer updates only from the Solver answer trajectories.
As shown in Table~\ref{prco_ablation}, removing Solver updates substantially reduces the overall improvement.
This is expected because the base model's reasoning is limited, and outcome-driven optimization on Solver trajectories is necessary for improving end-task accuracy.
Notably, PRCO w/o Solver still outperforms the baseline, indicating that utility-driven caption learning alone can improve final-answer accuracy and suggesting that the perception side remains a key bottleneck.
In contrast, removing Observer updates yields consistent drops across model scales and benchmarks.
This confirms that utility-driven evidence extraction provides complementary benefits on top of outcome-optimized reasoning by strengthening the intermediate evidence signal available to the Solver.
We also report the training curves in Fig.~\ref{fig:ablation_training_curves}, which show that PRCO achieves higher rewards throughout training.

\paragraph{Effect of caption-first warmup.}
We further ablate the caption-first warmup, where the Solver is first trained without image inputs to encourage reliance on the Observer caption before restoring full multimodal inputs.
As shown in Table~\ref{prco_ablation}, removing warmup degrades performance on both backbones, reducing the overall average on Qwen2.5-VL-7B from $49.63$ to $47.16$.
This suggests that warmup is important for encouraging caption usage. Without warmup, the Solver can rely on raw visual inputs too early, which weakens the learning signal for the Observer.
To further diagnose this behavior, we analyze how the standard deviation of caption rewards evolves during training in Fig.~\ref{fig:ablation_std}. Without warmup, the standard deviation decreases rapidly. This is consistent with weaker observed contrast among caption utilities, reducing the relative learning signal available to the Observer.

\subsection{More Results and Analysis}

\begin{figure}[t]
    \centering
    \includegraphics[width=\columnwidth]{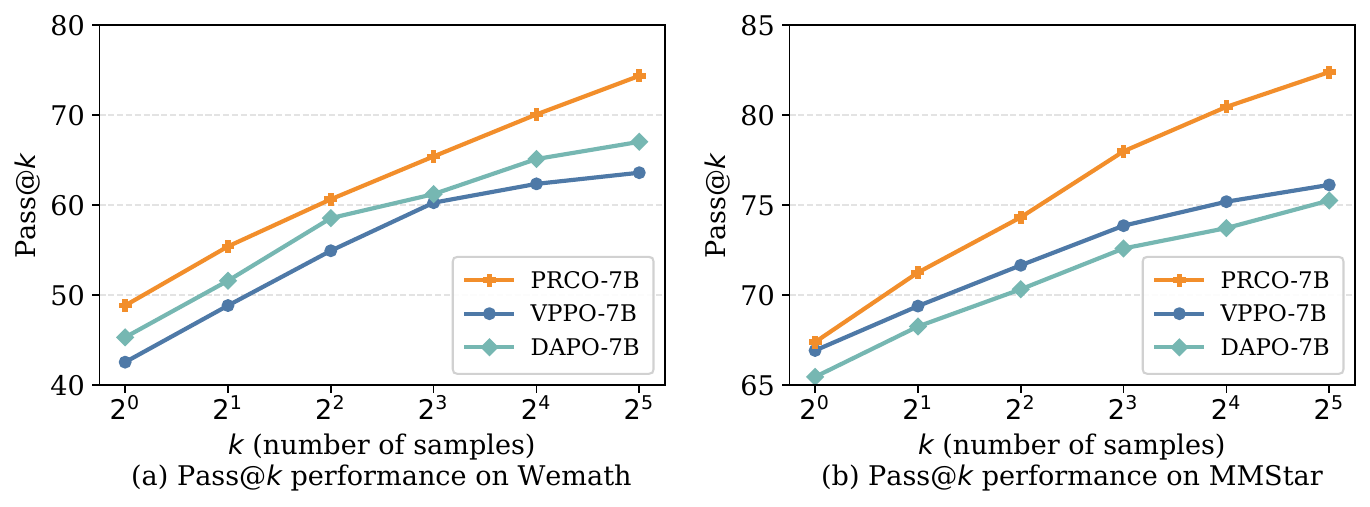}
    \caption{Pass@$k$ comparison on WeMath and MMStar for PRCO-7B, DAPO-7B, and VPPO-7B under different inference-time sampling budgets.}
    \label{passk}
\end{figure}

\paragraph{Pass@k Performance.}
Pass@k estimates the probability of obtaining at least one correct answer within $k$ attempts~\cite{chen2021evaluating}.
We compare PRCO-7B with two competitive Qwen2.5-VL-7B baselines, DAPO-7B and VPPO-7B, by estimating pass@k with \(k\in\{1,2,4,8,16,32\}\) sampled solutions per question.
We report the pass@k on WeMath and MMStar in Fig.~\ref{passk}.
As $k$ increases, PRCO-7B exhibits larger gains over the baselines on both benchmarks.
On WeMath, the gap over DAPO-7B grows from 3.53 at pass@1 to 7.33 at pass@32.
On MMStar, PRCO-7B is comparable to VPPO-7B at pass@1, while the margin increases from 0.47 at pass@1 to 6.27 at pass@32.
This trend suggests that PRCO-7B scales better with the sampling budget, indicating more robust reasoning capability.

\begin{figure}[t]
    \centering
    \includegraphics[width=\columnwidth]{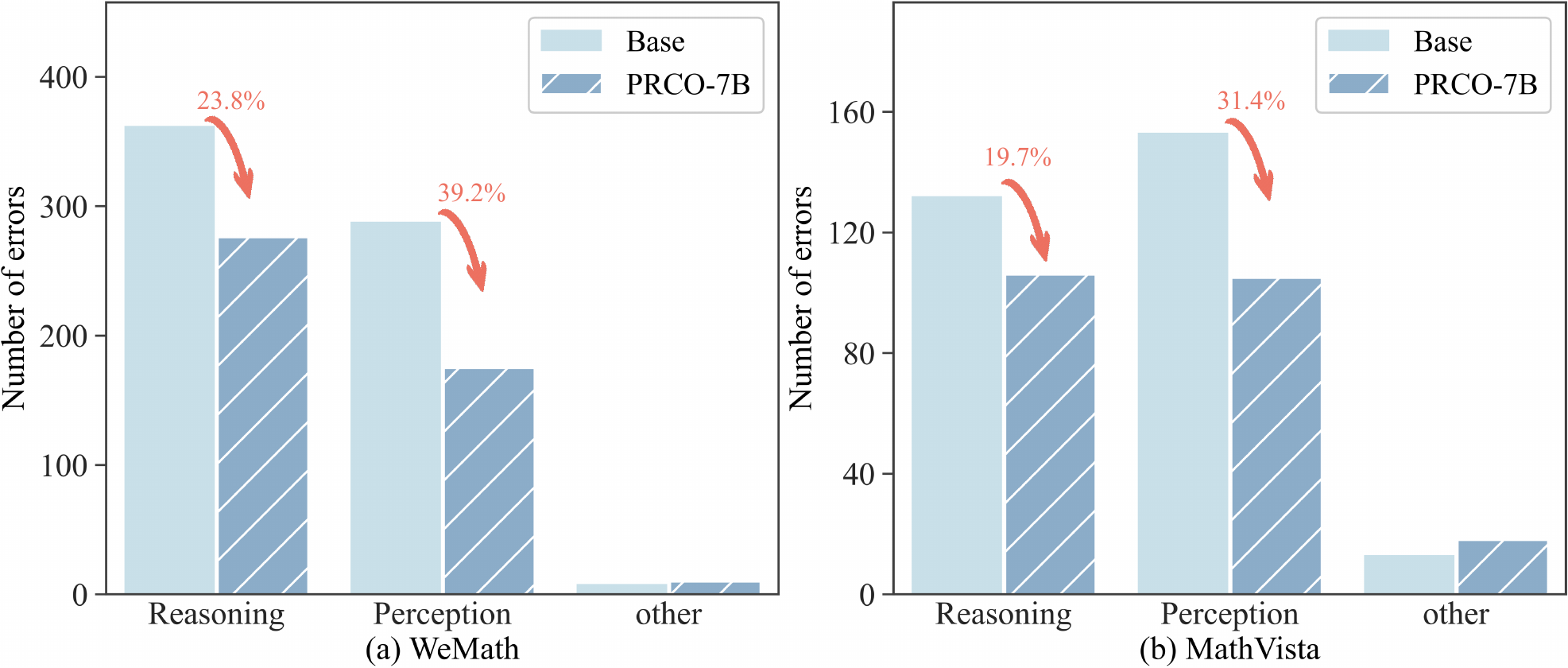}
    \caption{Error category analysis on WeMath and MathVista. Compared with Qwen2.5-VL-7B, PRCO-7B reduces both perception and reasoning errors. For presentation clarity, Knowledge and Extraction errors are grouped into the Other category.}
    \label{class_prco}
\end{figure}

\paragraph{Error category analysis.} We conduct an error-category analysis of Qwen2.5-VL-7B and PRCO-7B on WeMath and MathVista. Using the prompt in Fig.~\ref{fig:error_type_classification_prompt}, we use OpenAI’s GPT-5.1 model to categorize each incorrect prediction into five types: perception, reasoning, knowledge, extraction, and other. For reporting, we merge the last three into other errors. Compared with Qwen2.5-VL-7B, PRCO reduces both perception and reasoning errors, as shown in Fig.~\ref{class_prco}. On WeMath, PRCO reduces perception errors by 39.2\% and reasoning errors by 23.8\%. Notably, PRCO achieves a larger reduction in perception errors than GRPO, consistent with the results in Fig.~\ref{fig:grpo_diagnostic}. A similar trend is observed on MathVista, where PRCO reduces both perception and reasoning errors. These results suggest that separate and reliable learning signals improve question-grounded visual perception and enable more robust reasoning under explicit evidence guidance.

\begin{figure}[t]
    \centering
    \includegraphics[width=\columnwidth]{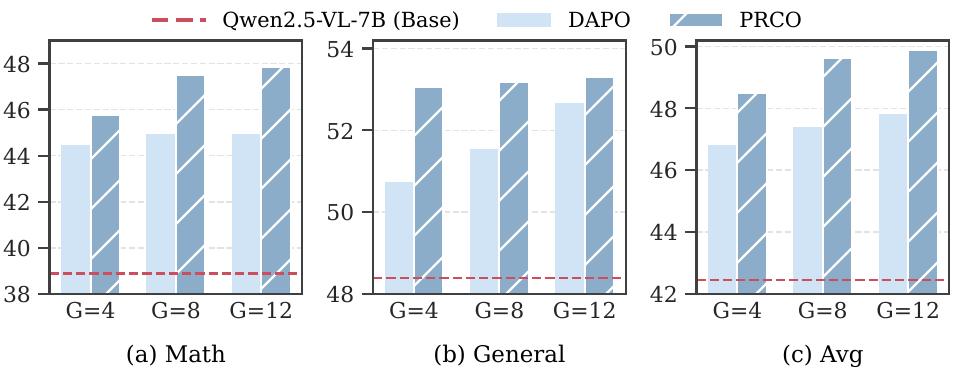}
    \caption{Effect of solver rollout group size $G_S$ on Qwen2.5-VL-7B. We vary $G_S$ among \{4, 8, 12\} and compare PRCO with DAPO on Math, General, and Avg. The dashed line denotes the base model performance.}
    \label{fig:solver_rollout_group_ablation}
\end{figure}

\paragraph{Effect of rollout group size.}
Rollout group size is a key hyperparameter in online RL, as it controls both the number of within-prompt samples and the training-time rollout budget.  We further study PRCO under different rollout budgets by first fixing the Observer group size and varying the Solver rollout group size $G_S$. As shown in Fig.~\ref{fig:solver_rollout_group_ablation}, increasing $G_S$ consistently improves both PRCO and DAPO, with a larger gain from 4 to 8 than from 8 to 12. Notably, PRCO with only $G_S{=}4$ already outperforms DAPO with $G{=}12$, underscoring the effectiveness of PRCO even with a much smaller Solver-side rollout group. These results support the benefit of role-specific learning signals under the reported sampling settings. We also study the effect of the Observer rollout group size $G_O$ in Appendix~\ref{app:more}.

\subsection{Case Study}

\begin{figure}[t]
    \centering
    \includegraphics[width=\columnwidth]{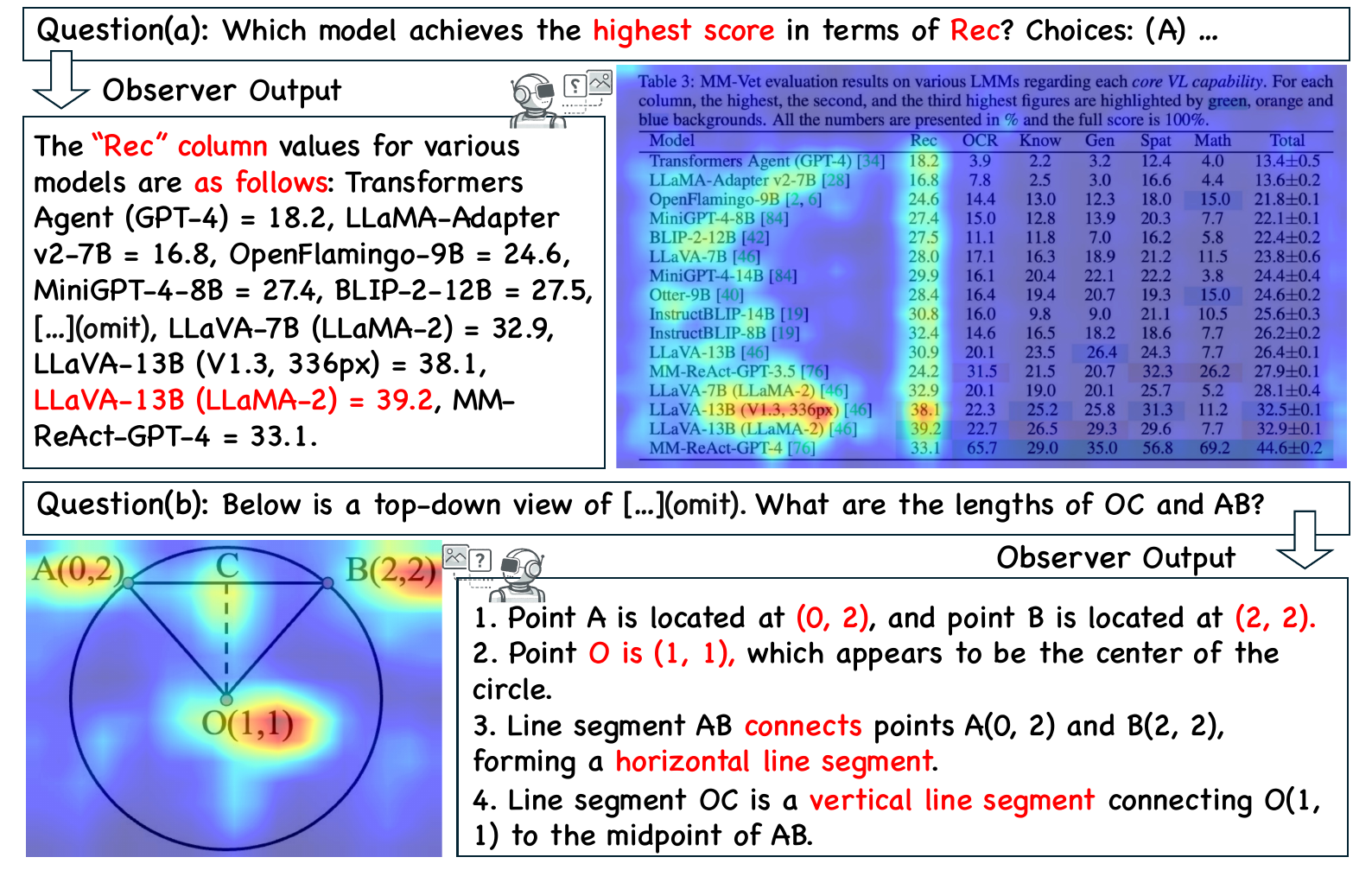}
    \caption{Qualitative analysis of PRCO-7B on two representative cases. For each case, we show the Observer output together with an attention overlay obtained by averaging attention to image tokens across all layers.}
    \label{fig:qualitative_observer_attention}
\end{figure}

To better understand how PRCO improves question-grounded visual perception, Fig.~\ref{fig:qualitative_observer_attention} presents two representative qualitative cases, showing the Observer outputs and the corresponding attention overlays obtained by averaging attention to image tokens across all layers. In case (a), the Observer accurately extracts only the question-relevant visual evidence from the \textit{Rec} column, rather than transcribing the full table, indicating that it preserves the evidence necessary for table-based option selection while avoiding unnecessary visual details. In case (b), the Observer localizes points, coordinates, and segment relations from the diagram, providing the full set of visual evidence required for solving the geometry question. In both cases, the attention overlays are concentrated on the corresponding question-relevant regions. More complete case studies are provided in Appendix~\ref{sec:case_studies}.

\section{Related Work}

\paragraph{RL with verifiable rewards for multimodal reasoning.}
Reinforcement learning with verifiable rewards (RLVR) defines rewards via automatic outcome verification.
It is often optimized with group-based PPO variants such as GRPO and DAPO \cite{shao2024deepseekmath,yu2025dapo}.
Recent work has begun to explore RLVR for MLLMs, with improvements in data construction, curricula, and rollout strategies.
Vision-R1 bootstraps multimodal chain-of-thought with staged RL schedules \citep{huang2025visionr1}, NoisyRollout perturbs images during rollouts to improve exploration and robustness \citep{liu2025noisyrollout}, and VL-Rethinker stabilizes training via selective replay and forced rethinking \citep{wang2025vlrethinker}.
RLVR has also been paired with explicit visual operations, e.g., Active-O3 \citep{zhu2025activeo3}, DeepEyes \citep{zheng2025deepeyes}, Pixel Reasoner \citep{su2025pixelreasoner}, and OpenThinkIMG \citep{su2025openthinkimg}.

\paragraph{Perception-aware RL for multimodal reasoning.}
Beyond outcome rewards, recent work incorporates perception-aware signals and objectives to improve visual perception in multimodal reasoning \citep{wang2025papo,xiao2025perceptionr1,zhang2025pearl}.
Perception-R1 introduces an explicit perception reward to score the fidelity of visual evidence in trajectories \citep{xiao2025perceptionr1}.
CapRL defines caption rewards by their question-answering utility for a vision-free LLM \citep{xing2025caprl}, while SOPHIA adopts semi-off-policy RL that propagates outcome rewards from external slow-thinking traces back to the model's visual understanding \citep{shen2025sophia}.
Other approaches use structured descriptions or caption-conditioned consistency regularization to improve downstream reasoning \citep{gou2025racro,tu2025cappo}.
PAPO integrates perception signals into policy optimization via objective-level regularization \citep{wang2025papo}.
Reward designs based on verifiable perception proxies or perception gates provide complementary supervision \citep{wang2025vicrit,zhang2025pearl}.
Credit assignment is further refined through visually dependent token reweighting~\citep{huang2025vppo}, step-level drawing-action rewards~\citep{huang2026sketchvl}, and verifier-guided updates on generated evidence spans in remote sensing~\citep{miao-etal-2026-grounding}.

\section{Conclusion}

In this paper, we presented PRCO, a dual-role RLVR framework for multimodal reasoning that disentangles perception and reasoning under a shared policy.
By assigning separate and reliable learning signals to an Observer for question-conditioned evidence captioning and a Solver for evidence-conditioned reasoning, PRCO enables perception-reasoning co-evolution during RLVR training.
Extensive experiments on eight challenging benchmarks showed consistent gains over strong RLVR baselines across model scales, while ablation and diagnostic analyses further validated the effectiveness of its key design choices. Overall, these results suggest that role-specific learning signals are a promising direction for improving multimodal reasoning under verifiable rewards.

\section*{Limitations}
Our current study focuses on multimodal reasoning benchmarks with concise and verifiable answers. Further evaluation is needed to determine how well PRCO generalizes to more open-ended generation settings. Extending the framework to broader multimodal generation tasks is an important direction for future work, since reward signals in these settings are often less well defined.
In addition, the Observer is trained with auxiliary supervision for leakage detection and answer verification. Although this auxiliary supervision is helpful in our setting, it may also introduce additional noise and computational overhead.
Finally, representing visual evidence as short captions is inherently lossy. Important aspects of the input, such as global structure (e.g., layout and texture), fine-grained spatial relations, and geometric details that are difficult to compress faithfully into text, may be only partially preserved. Future work could therefore explore richer intermediate representations for visual inputs that cannot be adequately captured by captions.

\section*{Ethical Considerations}
This work aims to improve multimodal reasoning by explicitly separating perception and reasoning during reinforcement learning. All experiments are conducted on publicly available datasets and benchmarks. As in prior work, these data sources may contain social biases, annotation artifacts, or other imperfections that can affect model behavior and evaluation outcomes. We do not identify additional ethical risks introduced specifically by our method beyond those already associated with multimodal model training and evaluation on existing public data. We encourage continued attention to data quality, transparent evaluation, and responsible reporting of model capabilities and limitations.
\section*{Acknowledgments}
This work was supported by the Shanghai Artificial Intelligence Laboratory. We thank the anonymous reviewers for their helpful feedback.
\bibliography{custom}

\appendix

\section{Additional Experiments and Analysis}
\label{sec:appendix}

\subsection{Evaluation Details}
\label{app:eval}
We evaluate our method on a diverse set of benchmarks spanning both math-related reasoning tasks and general multimodal tasks. Table~\ref{tab:benchmark} summarizes the benchmarks used in our evaluation, where the evaluation splits and reported metrics follow the settings in VLMEvalKit~\citep{duan2024vlmevalkit}. During evaluation, we strictly use the official prompts for all open-source MLLM baselines to avoid potential evaluation discrepancies. For PRCO, we use role-specific prompts for the Observer and Solver during inference. The Observer is prompted to produce a question-conditioned evidence caption, while the Solver is prompted to answer the question based on the caption and image.

\textbf{Math-Related Reasoning Tasks.} This category evaluates mathematical reasoning abilities.
\begin{itemize}
    \item \textbf{MathVerse}~\citep{zhang2024mathverse} is a benchmark for multimodal mathematical reasoning that examines whether MLLMs truly understand diagrams. By presenting each problem in multiple versions with different distributions of textual and visual information, it enables fine-grained analysis of a model's reliance on visual versus textual cues.
    \item \textbf{MathVision}~\citep{wang2024mathvision} focuses on competition-level multimodal math reasoning. Its problems are drawn from real mathematics competitions and cover multiple disciplines and difficulty levels, providing a challenging testbed for advanced reasoning over diagrams and symbolic content.
    \item \textbf{MathVista}~\cite{lu2023mathvista} is a comprehensive benchmark for visual mathematical reasoning. It covers diverse task types such as geometry, charts, tables, and scientific figures, making it a broad benchmark for evaluating mathematical reasoning in visually grounded settings.
    \item \textbf{WeMath}~\cite{qiao2024wemath} introduces a diagnostic evaluation paradigm for multimodal math reasoning. By decomposing problems into sub-problems based on knowledge concepts, it supports fine-grained analysis of a model's strengths and weaknesses.
    \item \textbf{DynaMath}~\cite{zou2024dynamath} is designed to evaluate the robustness and generalization of multimodal mathematical reasoning. It generates dynamic variations of seed problems, allowing evaluation of whether a model can maintain consistent reasoning under controlled changes.
\end{itemize}

\textbf{General Multimodal Tasks.} This category evaluates broader multimodal understanding abilities.
\begin{itemize}
    \item \textbf{LogicVista}~\cite{xiao2024logicvista} focuses on logical reasoning in visual contexts. Although not limited to mathematics, it is useful for evaluating whether models can perform structured reasoning grounded in diagrams and other visual inputs.
    \item \textbf{MMMU-Pro}~\cite{yue2024mmmupro} is an enhanced benchmark for multidisciplinary multimodal understanding and reasoning. It is designed to reduce shortcuts from textual clues and provide a more rigorous evaluation of genuine visual understanding across subjects.
    \item \textbf{MMStar}~\cite{chen2024mmstar} is a curated benchmark for core multimodal reasoning abilities. Its samples are designed to require genuine visual understanding, making it a concise but challenging benchmark for multimodal reasoning evaluation.
\end{itemize}

\textbf{Evaluation parameters.} Unless otherwise specified, we use greedy decoding for single-sample evaluation, with temperature set to 0.0, top-$p$ to 1.0, top-$k$ to -1, and the maximum number of generated tokens to 2048. For pass@$k$ evaluation, we instead use temperature 0.6, top-$p$ 0.95, and top-$k$ -1.

\textbf{Implementation details of error analysis.} For error categorization, we use OpenAI’s GPT-5.1 with temperature set to 0.0. For each incorrect prediction, the classifier receives the following inputs simultaneously: Image, Question, Model response, and Gold answer. The detailed classification prompt is shown in Fig.~\ref{fig:error_type_classification_prompt}. We classify each error into one of five categories: Perception, Reasoning, Knowledge, Extraction, and Other. In practice, we find that the numbers of Knowledge, Extraction, and Other errors are relatively small. Therefore, for clearer visualization, we merge these three categories into a single Other category in Fig.~\ref{class_prco}.

\begin{table*}[t]
\centering
\resizebox{\textwidth}{!}{%
\begin{tabular}{lccc}
\toprule
\textbf{Benchmark} & \textbf{Evaluation Split} & \textbf{Num. of Samples} & \textbf{Metric} \\
\midrule
MathVerse~\citep{zhang2024mathverse}  & MathVerse\_MINI\_Vision\_Only & 788 & Overall \\
MathVista~\citep{lu2023mathvista}  & MathVista\_MINI               & 1000 & acc \\
MathVision~\citep{wang2024mathvision} & MathVision                    & 3040 & acc \\
WeMath~\citep{qiao2024wemath}     & WeMath                        & 1740 & Score (Strict) \\
DynaMath~\citep{zou2024dynamath}   & DynaMath   & 5010 & Overall (Worst Case) \\
LogicVista~\citep{xiao2024logicvista} & LogicVista                    & 448  & Overall \\
MMMU-Pro~\citep{yue2024mmmupro}   & MMMU\_Pro\_V                  & 1730 & Overall \\
MMStar~\citep{chen2024mmstar}     & MMStar                        & 1500 & Overall \\
\bottomrule
\end{tabular}%
}
\caption{Details of the benchmarks we evaluate. The evaluation splits and reported metrics follow the settings in VLMEvalKit~\citep{duan2024vlmevalkit}. We report single-sample greedy scores under each benchmark’s official VLMEvalKit metric, which we denote as accuracy for simplicity.}
\label{tab:benchmark}
\end{table*}

\subsection{Training Details}
\label{app:train}
In this section, we describe the training details of the different methods.  All training is conducted on 8 NVIDIA H200 GPUs.

For the RLVR baselines GRPO and DAPO, we follow the EasyR1 implementations~\citep{zheng2025easyr1} exactly. GRPO uses clipping factors $\epsilon_l=0.2$ and $\epsilon_h=0.3$ with a reference KL penalty coefficient $\beta=0.01$, while DAPO uses $\epsilon_l=0.2$ and $\epsilon_h=0.28$, removes the reference KL term, enables token-level loss averaging, and adopts dynamic sampling with a maximum of 20 retries. Other training hyperparameters, including the number of training steps, rollout batch size, and maximum sequence length, are summarized in Table~\ref{tab:training_hyperparameters}. More implementation details can be found in the EasyR1 codebase.

Our implementation of PRCO is based on the EasyR1 framework~\citep{zheng2025easyr1}. We train all models on the ViRL39K and use MMK12~\citep{meng2025mmeureka} as the validation set. Following DAPO~\citep{yu2025dapo}, we use dynamic sampling, clip-higher, and token-level policy gradient loss. The clipping factors are set to $\epsilon_l=0.2$ and $\epsilon_h=0.28$, respectively, and no KL-divergence penalty is applied.
We also remove the standard-deviation normalization term when computing the grouped advantage in PRCO. We find this design more suitable for role-specific optimization, as it preserves the original relative reward differences within each role and leads to more faithful advantage updates for both the Observer and the Solver. Table~\ref{tab:training_hyperparameters} summarizes the main hyperparameters used in our experiments. For PRCO, the maximum rollout length is set to 1024 tokens for the Observer and 2048 tokens for the Solver.
We use Qwen3-VL-8B-Instruct~\citep{yang2025qwen3} as the auxiliary model for answer leakage checking.
We also adopt a caption-first warmup for the first 40 training steps, during which the Solver is trained without image inputs to encourage caption conditioning before restoring full multimodal inputs.

\begin{table*}[htbp]
\centering
\small
\renewcommand{\arraystretch}{1.1}
\resizebox{\textwidth}{!}{
\begin{tabular}{lcccccccccc}
\toprule
\textbf{Method} & \textbf{lr} & \textbf{Max Len.} & \textbf{Steps} & \textbf{Warmup} & \textbf{Opt.} & \textbf{Rollout BS} & \textbf{Freeze VT} & \textbf{Temp.} & \textbf{top-$p$} & \textbf{top-$k$} \\
\midrule
GRPO & 1e-6 & 2048 & 200 & -- & AdamW & 384 & False & 1 & 1.0 & -1 \\
DAPO & 1e-6 & 2048 & 200 & -- & AdamW & 384 & False & 1 & 1.0 & -1 \\
PRCO & 1e-6 & 1024 / 2048 & 200 & 40 & AdamW & 384 & False & 1 & 1.0 & -1 \\
\bottomrule
\end{tabular}
}
\caption{Training hyperparameters used in our experiments. For GRPO and DAPO, Max Len. denotes the maximum rollout length of the single policy. For PRCO, it denotes Observer / Solver maximum rollout lengths.}
\label{tab:training_hyperparameters}
\end{table*}

\subsection{Prompt Templates}
In this section, we present the prompts used in our experiments. For the RLVR baselines, including GRPO and DAPO, we follow the prompt setting used in EasyR1 \citep{zheng2025easyr1}, where the model is asked to first reason through the problem and then provide the final answer in a boxed format. For PRCO, we use role-specific prompts for the Observer and Solver. The Observer is prompted to produce a question-conditioned evidence caption that captures the question-relevant visual evidence without revealing the final answer, while the Solver is prompted to answer the question based primarily on the caption and consult the image only when necessary. Fig.~\ref{fig:observer_prompt} shows the prompts used for PRCO and the RLVR baselines. Beyond the main training and inference prompts, we also employ auxiliary prompts for both training and analysis. Specifically, we use a leakage-checking prompt to verify that the Observer caption does not directly reveal the answer, and an error-type classification prompt to categorize model failures in the error analysis. Fig.~\ref{fig:error_type_classification_prompt} shows these auxiliary prompts.

\subsection{More Results and Analysis}
\label{app:more}
We additionally evaluate three PRCO-7B variants to study the roles of Solver-side visual grounding, leakage suppression, and co-evolving utility feedback: (\textit{i}) PRCO w/ $I^S=\emptyset$, which keeps the Solver image input empty throughout the RL stage; (\textit{ii}) PRCO w/o Leakage Checker, which removes the leakage checker from the Observer utility reward; and (\textit{iii}) PRCO w/ Fixed Utility Estimator, which replaces the co-evolving Solver with a fixed Qwen2.5-VL-7B for caption utility estimation.

Table~\ref{tab:prco_more_results_7b} shows that all three variants underperform the full PRCO-7B, confirming that PRCO's gains arise from the combination of evidence-conditioned reasoning, utility reward with leakage checking, and Observer--Solver co-evolution.
The largest drop is observed for PRCO w/ $I^S=\emptyset$, where the Solver never regains access to the image after the caption-first warmup. This suggests that while restricting the Solver to caption-based evidence is beneficial in early training, access to the image remains important in later RL optimization. In PRCO, the caption is the primary evidence channel, but restored image access still helps recover global structure, fine-grained spatial relations, and geometric details that are difficult to fully compress into text. 
Removing the leakage checker also degrades the overall average, indicating that suppressing answer leakage is important for learning useful intermediate evidence. Without leakage checking, the Observer is more likely to exploit answer shortcutting by placing the final answer directly in the caption, rather than extracting question-relevant visual evidence. This weakens the utility reward as a learning signal for evidence extraction and blurs the credit assignment between perception and reasoning.
PRCO w/ Fixed Utility Estimator further underperforms standard PRCO, suggesting that Observer learning benefits more from utility feedback that co-evolves with the Solver and remains better aligned with its changing information needs.

\begin{table*}[t]
\centering
\setlength{\tabcolsep}{4.5pt}
\renewcommand{\arraystretch}{1.05}
\resizebox{\textwidth}{!}{%
\begin{tabular}{lccccccccc}
\toprule
Setting & MathVerse & MathVision & MathVista & WeMath & DynaMath & LogicVista & MMMU-Pro & MMStar & Avg. \\
\midrule
Qwen2.5-VL-7B (Base) & 43.02 & 25.46 & 70.20 & 35.43 & 20.35 & 45.41 & 35.49 & 64.26 & 42.45 \\
\midrule
\textbf{PRCO-7B (Ours)} & 49.49 & 30.86 & 77.10 & 50.29 & 29.74 & 49.66 & 42.08 & 67.80 & 49.63 \\
PRCO w/ $I^S=\emptyset$ & 47.34 & 28.82 & 75.20 & 43.81 & 28.94 & 48.77 & 42.89 & 66.13 & 47.74 {\color{red}\scriptsize$\downarrow$1.89} \\
PRCO w/ Fixed Utility Estimator & 49.24 & 29.67 & 76.70 & 45.52 & 29.74 & 48.55 & 40.17 & 66.20 & 48.22 {\color{red}\scriptsize$\downarrow$1.41} \\
PRCO w/o Leakage Checker & 50.00 & 29.31 & 76.00 & 44.19 & 29.54 & 51.45 & 41.33 & 67.60 & 48.68 {\color{red}\scriptsize$\downarrow$0.95} \\
\bottomrule
\end{tabular}%
}
\caption{Additional ablations of PRCO-7B. We report benchmark scores on eight benchmarks. \textcolor{red}{Red} downward arrows in the Avg. column indicate the drop relative to PRCO-7B (ours).}
\label{tab:prco_more_results_7b}
\end{table*}

\begin{figure}[t]
    \centering
    \includegraphics[width=\columnwidth]{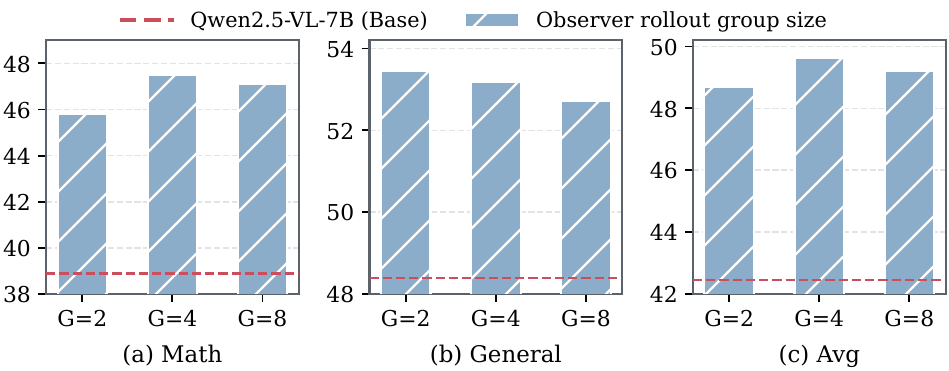}
    \caption{Ablation of the observer rollout group size $G_O$ in PRCO on Qwen2.5-VL-7B. Bars show different $G_O$ settings ($2,4,8$) on Math, General, and Avg, and the dashed line indicates the base model performance.}
    \label{fig:Observer_rollout_group_ablation}
\end{figure}

\begin{figure}[t]
    \centering
    \includegraphics[width=\columnwidth]{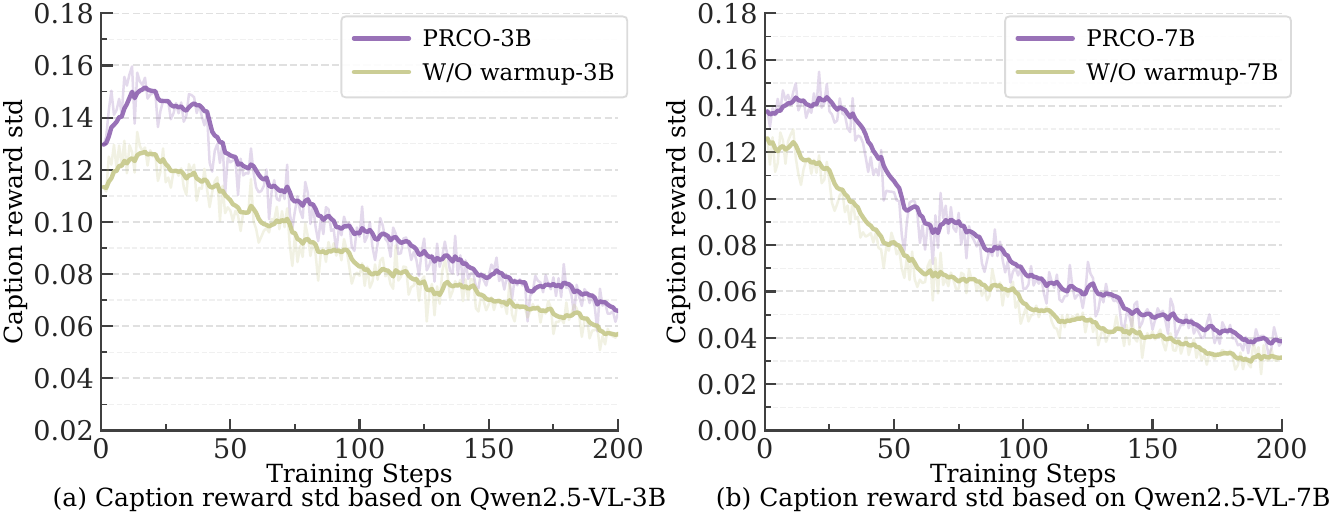}
    \caption{Training caption reward standard deviation curves of PRCO and its W/O warmup variant with Qwen2.5-VL-3B and Qwen2.5-VL-7B as backbones.}
    \label{fig:ablation_std}
\end{figure}

\paragraph{Observer rollout group size.}
We further vary the observer rollout group size \(G_O\) on Qwen2.5-VL-7B while keeping the Solver rollout group size fixed. As shown in Fig.~\ref{fig:Observer_rollout_group_ablation}, the effect of \(G_O\) is not monotonic: performance improves from \(G_O=2\) to \(G_O=4\), but slightly declines at \(G_O=8\). A possible explanation is that perception saturates earlier than reasoning, which is also consistent with Fig.~\ref{fig:ablation_training_curves}, where the variant without Solver updates reaches its plateau relatively early.
Once the Observer already provides sufficiently informative captions, further increasing \(G_O\) yields diminishing returns and may reduce the relative benefit of allocating more rollouts to the Solver.
At the reported settings, \(G_O=8, G_S=8\) performs worse than \(G_O=4, G_S=12\), as shown in Fig.~\ref{fig:Observer_rollout_group_ablation} and Fig.~\ref{fig:solver_rollout_group_ablation}. These settings vary both the caption and answer counts and do not constitute an equal-compute comparison. We use \(G_O=4\) by default.

\paragraph{Reliability of the Observer utility reward.}
To examine whether the utility signal agrees with an external assessment of caption quality, we sample 200 ViRL39K training questions with four candidate captions each and use GPT-5.5 to rate informativeness from 1 (erroneous or irrelevant) to 4 (accurate and comprehensive), given the image, question, and caption. These ratings yield a within-question Spearman correlation of 0.51 with Observer rewards and a pairwise ranking accuracy of 73.5\%. This moderately strong agreement supports the utility signal, although model-based ratings remain an empirical proxy rather than ground truth.

\subsection{Discussion of Utility Feedback and Role-wise Updates}
\label{app:utility_analysis}

\paragraph{Interpreting the utility reward.}
The Observer reward in Eq.~\eqref{eq:observer_reward} evaluates a caption through the current Solver's success, subject to leakage checking. Averaging independent responses from a fixed Solver policy reduces sampling noise, while the expected reward depends on the Solver's ability to use the evidence. Useful evidence may receive a low reward when the Solver cannot exploit it, and image access may allow success despite an incomplete caption. The reward thus provides task-oriented feedback without directly verifying every visual statement. The reward-reliability analysis complements this signal by examining its agreement with an external assessment of caption quality. Together with the leakage-checker ablation, it supports the use of downstream utility alongside explicit constraints against answer shortcutting.

\paragraph{Complementarity of role-wise updates.}
Table~\ref{prco_ablation} distinguishes updates from evidence captions and final-answer trajectories. At 7B, removing Observer updates lowers the average by $1.44$ points, while removing Solver updates lowers it by $5.11$ points. The larger drop without Solver updates highlights the importance of outcome-driven reasoning optimization. The remaining gap without Observer updates indicates that Solver optimization alone does not recover the full benefit of learning task-oriented evidence. These differences describe the reported ablations and are not additive contributions to the total gain. Since both roles share parameters, either update can affect both behaviors; separate comparison groups retain role-specific feedback within this shared model.

\subsection{PRCO on Qwen3-VL-8B-Instruct}
\label{app:qwen3}
To further evaluate PRCO on a stronger vision-language backbone, we also train Qwen3-VL-8B-Instruct~\citep{yang2025qwen3}. The training details are exactly the same as those in Appendix~\ref{app:train}. On this backbone, we compare PRCO with two RLVR baselines, DAPO and GRPO. As shown in Table~\ref{tab:qwen3}, PRCO outperforms both GRPO and DAPO on Qwen3-VL-8B-Instruct, further demonstrating its effectiveness on stronger models.

\begin{table*}[t]
\centering
\setlength{\tabcolsep}{4.0pt}
\renewcommand{\arraystretch}{1.05}
\resizebox{\textwidth}{!}{%
\begin{tabular}{lccccccccc}
\toprule
Setting & MathVerse (Vision Only) & MathVision & MathVista & WeMath & DynaMath & LogicVista & MMMU-Pro-V & MMStar & Avg. \\
\midrule
Qwen3-VL-8B-Instruct (Base) & 54.57 & 36.64 & 74.90 & 51.33 & 38.12 & 53.02 & 41.27 & 69.40 & 52.41 \\
\midrule
GRPO & 64.59 & 46.94 & \underline{78.30} & 59.62 & 41.12 & \underline{61.75} & 52.72 & 73.00 & 59.75 \\
DAPO & \underline{65.61} & \underline{50.13} & \underline{78.30} & \underline{66.86} & \underline{43.31} & \textbf{62.19} & \underline{53.87} & \underline{74.07} & \underline{61.79} \\
\textbf{PRCO (Ours)} & \textbf{69.67} & \textbf{51.78} & \textbf{79.00} & \textbf{68.57} & \textbf{46.11} & 60.40 & \textbf{54.57} & \textbf{74.27} & \textbf{63.05} \\
\bottomrule
\end{tabular}%
}
\caption{Comparison of PRCO with GRPO and DAPO on Qwen3-VL-8B-Instruct. We report benchmark scores on eight benchmarks. The best and second-best results within each backbone are highlighted in \textbf{bold} and \underline{underlined}.}
\label{tab:qwen3}
\end{table*}

\subsection{Case Study}
\label{sec:case_studies}
For Fig.~\ref{fig:qualitative_observer_attention}, we construct the attention heatmap by extracting attention weights from the Observer’s generated output tokens to visual tokens, averaging them across all heads and transformer layers, and mapping the aggregated scores back to the 2D visual-token layout.

\begin{figure*}[t]
\centering
\begin{casebox}{1:}

\noindent
\begin{minipage}{0.35\linewidth}
    \includegraphics[width=\linewidth]{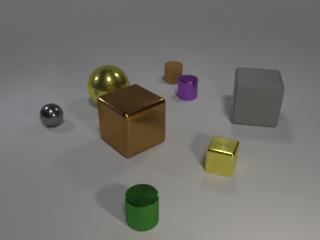}
\end{minipage}
\hfill
\begin{minipage}{0.62\linewidth}
    \centering
    {\textbf{Problem: } Subtract all big gray matte things. Subtract all small metallic cylinders. How many objects are left?}
\end{minipage}

\par\medskip

\noindent
\begin{minipage}{\linewidth}
\small

{\bfseries Observer Output.}

\begin{itemize}[leftmargin=1.3em, itemsep=2pt, topsep=3pt, parsep=0pt]
    \item A large golden metallic cube is present in the center-left.
    \item A large yellow metallic sphere is located towards the back-left.
    \item A small metallic gray sphere is on the far left.
    \item A small yellow metallic cube is near the center-right.
    \item A large gray matte cube is on the far right.
    \item A small green metallic cylinder is in the foreground.
    \item A small purple metallic cylinder is near the center.
    \item A small brown matte cylinder is in the background.
\end{itemize}

\par\smallskip
\hrule
\par\smallskip

{\bfseries Solver Output.}

\texttt{<think>}

To determine the number of objects left after the specified subtractions, we start by identifying all objects in the image.

First, we subtract all big gray matte things. There is one big gray matte cube, so we remove it.

Remaining objects:
\begin{itemize}[leftmargin=1.3em, itemsep=2pt, topsep=3pt, parsep=0pt]
    \item Large yellow metallic sphere.
    \item Large metallic gold cube.
    \item Small metallic gray sphere.
    \item Small brown matte cylinder.
    \item Small purple metallic cylinder.
    \item Small green metallic cylinder.
    \item Small yellow metallic cube.
\end{itemize}

Next, we subtract all small metallic cylinders. There are two small metallic cylinders: the purple one and the green one. Removing these:

Remaining objects:
\begin{itemize}[leftmargin=1.3em, itemsep=2pt, topsep=3pt, parsep=0pt]
    \item Large yellow metallic sphere.
    \item Large metallic gold cube.
    \item Small metallic gray sphere.
    \item Small brown matte cylinder.
    \item Small yellow metallic cube.
\end{itemize}

Thus, the number of objects left is 5.

\texttt{</think>}
\par\medskip
\noindent $\boxed{5}$ \hspace{5pt}$\leftarrow$ \textcolor{mygreen}{\texttt{Correct}}

\end{minipage}
\end{casebox}
\caption{Qualitative example of PRCO-7B on synthetic object filtering. The Observer enumerates the relevant objects together with size, color, material, shape, and coarse spatial cues, converting the scene into a question-conditioned object inventory. PRCO performs discrete filtering and counting from explicit visual evidence.}
\label{fig:case_object_filter}
\end{figure*}

\begin{figure*}[t]
\centering
\begin{casebox}{2:}

\noindent
\begin{minipage}{0.35\linewidth}
    \includegraphics[width=\linewidth]{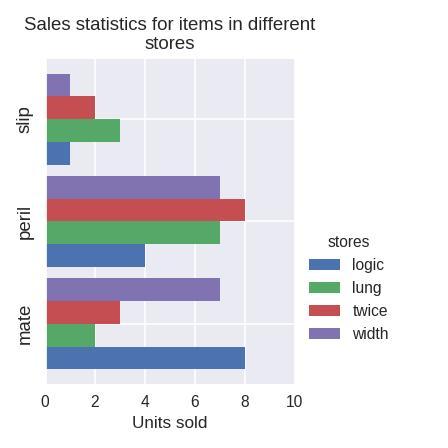}
\end{minipage}
\hfill
\begin{minipage}{0.62\linewidth}
    \centering
    {\textbf{Problem: } How many items sold less than 1 units in at least one store?}
\end{minipage}

\par\medskip

\noindent
\begin{minipage}{\linewidth}
\small

{\bfseries Observer Output.}

\begin{itemize}[leftmargin=1.3em, itemsep=2pt, topsep=3pt, parsep=0pt]
    \item For the item ``slip'', the sales for ``logic'' are 1 unit, for ``lung'' are 3 units, for ``twice'' are 2 units, and for ``width'' are 1 unit.
    \item For the item ``peril'', the sales for ``logic'' are 4 units, for ``lung'' are 7 units, for ``twice'' are 8 units, and for ``width'' are 7 units.
    \item For the item ``mate'', the sales for ``logic'' are 8 units, for ``lung'' are 2 units, for ``twice'' are 3 units, and for ``width'' are 7 units.
\end{itemize}

\par\smallskip
\hrule
\par\smallskip

{\bfseries Solver Output.}

\texttt{<think>}

To determine how many items sold less than 1 unit in at least one store, we need to carefully examine the bar lengths for each item in the given chart.

\begin{enumerate}[leftmargin=1.5em, itemsep=3pt, topsep=3pt, parsep=0pt]
    \item \textbf{Item ``slip'':}
    \begin{itemize}[leftmargin=1.3em, itemsep=1pt, topsep=2pt, parsep=0pt]
        \item Logic: 1 unit (not less than 1).
        \item Lung: 3 units (not less than 1).
        \item Twice: 2 units (not less than 1).
        \item Width: 1 unit (not less than 1).
    \end{itemize}
    Since all bars for ``slip'' are either 1 or greater, none of them are less than 1.

    \item \textbf{Item ``peril'':}
    \begin{itemize}[leftmargin=1.3em, itemsep=1pt, topsep=2pt, parsep=0pt]
        \item Logic: 4 units (not less than 1).
        \item Lung: 7 units (not less than 1).
        \item Twice: 8 units (not less than 1).
        \item Width: 7 units (not less than 1).
    \end{itemize}
    All bars for ``peril'' are greater than or equal to 1, so none are less than 1.

    \item \textbf{Item ``mate'':}
    \begin{itemize}[leftmargin=1.3em, itemsep=1pt, topsep=2pt, parsep=0pt]
        \item Logic: 8 units (not less than 1).
        \item Lung: 2 units (not less than 1).
        \item Twice: 3 units (not less than 1).
        \item Width: 7 units (not less than 1).
    \end{itemize}
    All bars for ``mate'' are greater than or equal to 1, so none are less than 1.
\end{enumerate}

Upon reviewing all items, none of the bars for any item show values less than 1.

Therefore, the number of items sold less than 1 unit in at least one store is \(0\).

\texttt{</think>}
\par\medskip
\noindent $\boxed{0}$ \hspace{5pt}$\leftarrow$ \textcolor{mygreen}{\texttt{Correct}}

\end{minipage}
\end{casebox}
\caption{Qualitative example of PRCO-7B on bar-chart threshold counting. The Observer transcribes item-wise values across stores into explicit textual evidence, and the Solver then checks the threshold condition and aggregates over items. PRCO supports accurate chart value extraction and evidence-conditioned counting through a compact intermediate representation.}
\label{fig:case_chart_count}
\end{figure*}

\begin{figure*}[t]
\centering
\begin{casebox}{3:}

\noindent
\begin{minipage}{0.45\linewidth}
    \includegraphics[width=\linewidth]{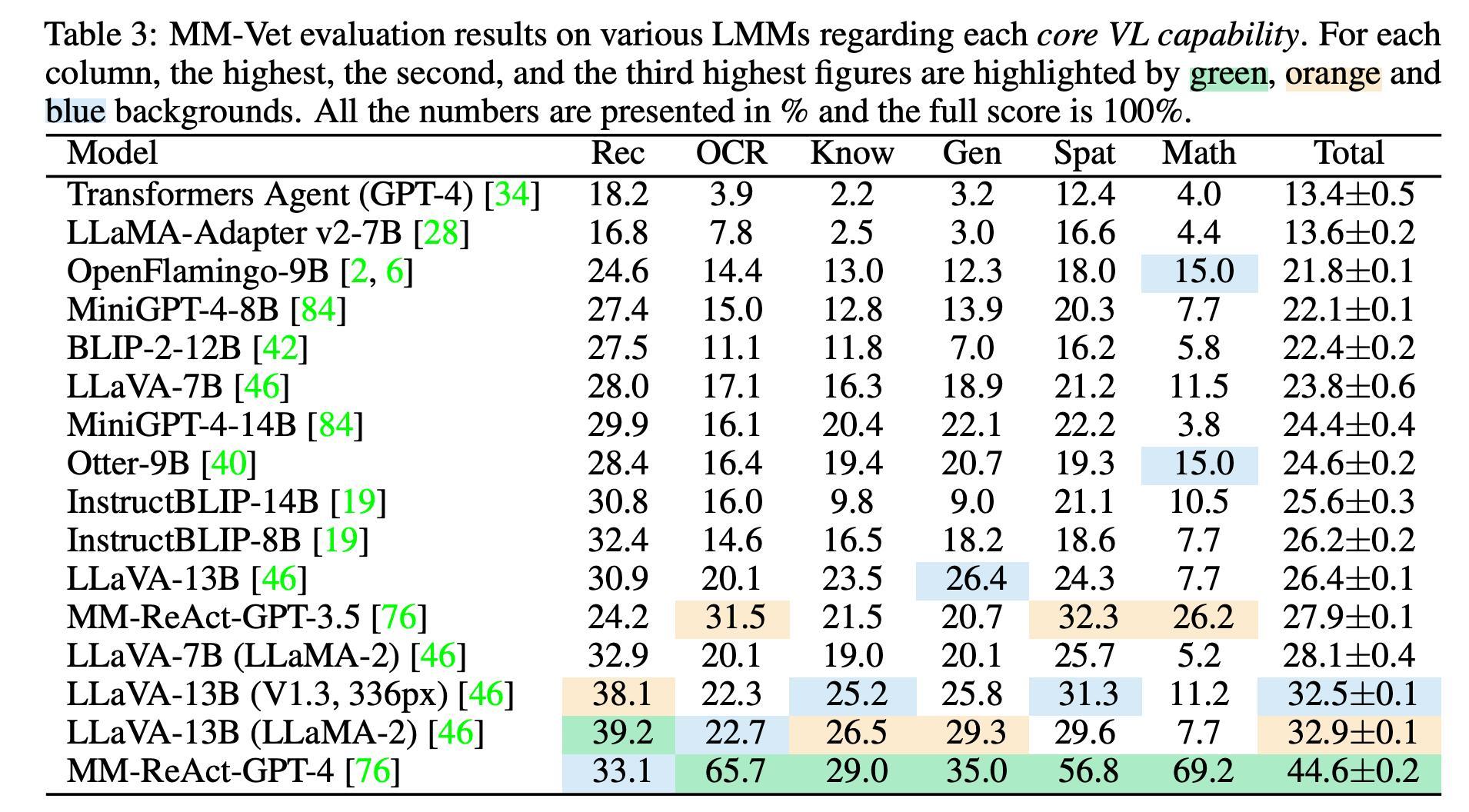}
\end{minipage}
\hfill
\begin{minipage}{0.52\linewidth}
    \small
    \centering
    {\textbf{Problem: } Which model achieves the highest score in terms of Rec?}
    Choices: (A) Transformers Agent (GPT-4), (B) LLaMA-Adapter v2-7B, (C) LLaVA-7B, (D) Otter-9B,\\
    (E) MM-ReAct-GPT-3.5, (F) LLaVA-13B (LLaMA-2), (G) MM-ReAct-GPT-4
\end{minipage}

\par\medskip

\noindent
\begin{minipage}{\linewidth}
\small

{\bfseries Observer Output.}

\begin{itemize}[leftmargin=1.3em, itemsep=2pt, topsep=3pt, parsep=0pt]
    \item The ``Rec'' column values for various models are as follows: Transformers Agent (GPT-4) = 18.2, LLaMA-Adapter v2-7B = 16.8, OpenFlamingo-9B = 24.6, MiniGPT-4-8B = 27.4, BLIP-2-12B = 27.5, LLaVA-7B = 28.0, MiniGPT-4-14B = 29.9, Otter-9B = 28.4, InstructBLIP-14B = 30.8, InstructBLIP-8B = 32.4, LLaVA-13B = 30.9, MM-ReAct-GPT-3.5 = 24.2, LLaVA-7B (LLaMA-2) = 32.9, LLaVA-13B (V1.3, 336px) = 38.1, LLaVA-13B (LLaMA-2) = 39.2, MM-ReAct-GPT-4 = 33.1.
\end{itemize}

\par\smallskip
\hrule
\par\smallskip

{\bfseries Solver Output.}

\texttt{<think>}

To determine which model achieves the highest score in terms of ``Rec'' (Recognition), we need to identify the highest value in the ``Rec'' column of the table.

The ``Rec'' scores for the models are as follows:
\begin{itemize}[leftmargin=1.3em, itemsep=1pt, topsep=2pt, parsep=0pt]
    \item Transformers Agent (GPT-4): 18.2
    \item LLaMA-Adapter v2-7B: 16.8
    \item OpenFlamingo-9B: 24.6
    \item MiniGPT-4-8B: 27.4
    \item BLIP-2-12B: 27.5
    \item LLaVA-7B: 28.0
    \item MiniGPT-4-14B: 29.9
    \item Otter-9B: 28.4
    \item InstructBLIP-14B: 30.8
    \item InstructBLIP-8B: 32.4
    \item LLaVA-13B: 30.9
    \item MM-ReAct-GPT-3.5: 24.2
    \item LLaVA-7B (LLaMA-2): 32.9
    \item LLaVA-13B (V1.3, 336px): 38.1
    \item LLaVA-13B (LLaMA-2): 39.2
    \item MM-ReAct-GPT-4: 33.1
\end{itemize}

Among these, the highest score in the ``Rec'' column is 39.2, which belongs to the model ``LLaVA-13B (LLaMA-2)''.

Therefore, the correct choice is (F).

\texttt{</think>}
\par\medskip
\noindent $\boxed{F}$ \hspace{5pt}$\leftarrow$ \textcolor{mygreen}{\texttt{Correct}}

\end{minipage}
\end{casebox}
\caption{Qualitative example of PRCO-7B on table-based option selection. The Observer extracts only the question-relevant entries in the \texttt{Rec} column, rather than transcribing the full table, after which the Solver identifies the maximum and maps it to the correct answer option. PRCO focuses on question-relevant evidence while avoiding unnecessary visual details.}
\label{fig:case_table_selection}
\end{figure*}

\begin{figure*}[t]
\centering
\begin{casebox}{4:}

\noindent
\begin{minipage}{0.3\linewidth}
    \includegraphics[width=\linewidth]{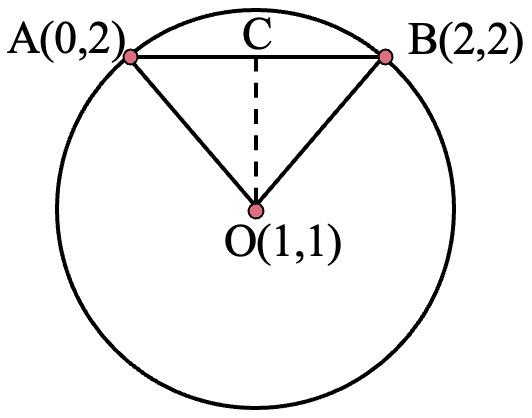}
\end{minipage}
\hfill
\begin{minipage}{0.65\linewidth}
    \small
    \centering
    {\textbf{Problem: } Below is a top-down view of a circular water surface. In the diagram, one unit length represents 1 meter. What is the degree measure of $\angle BOA$? What are the lengths of $OC$ and $AB$?}\\
    Options: A. $30^\circ$, the lengths of $OC$ and $AB$ are 1m and 2m respectively;\\
    B. $45^\circ$, the lengths of $OC$ and $AB$ are 2m and 2m respectively;\\
    C. $90^\circ$, the lengths of $OC$ and $AB$ are 1m and 2m respectively;\\
    D. Cannot be determined; E. No correct answer
\end{minipage}

\par\medskip

\noindent
\begin{minipage}{\linewidth}
\small

{\bfseries Observer Output.}
\begin{itemize}[leftmargin=1.3em, itemsep=2pt, topsep=3pt, parsep=0pt]
    \item Point A is located at $(0,2)$, and point B is located at $(2,2)$.
    \item Point O is at $(1,1)$, which appears to be the center of the circle.
    \item Line segment AB connects points A$(0,2)$ and B$(2,2)$, forming a horizontal line segment.
    \item Line segment OC is a vertical line segment connecting O$(1,1)$ to the midpoint of AB.
\end{itemize}

\par\smallskip
\hrule
\par\smallskip

{\bfseries Solver Output.}

\texttt{<think>}

To determine the degree measure of $\angle BOA$ and the lengths of $OC$ and $AB$, we start by finding the radius of the circle. Given points $A(0,2)$ and $B(2,2)$, the center $O(1,1)$ of the circle can be used to find the radius.

First, calculate the length $OA$:
\[
OA=\sqrt{(1-0)^2+(1-2)^2}=\sqrt{1^2+(-1)^2}=\sqrt{1+1}=\sqrt{2}.
\]

Similarly, calculate $OB$:
\[
OB=\sqrt{(1-2)^2+(1-2)^2}=\sqrt{(-1)^2+(-1)^2}=\sqrt{1+1}=\sqrt{2}.
\]

Since $OA=OB=\sqrt{2}$, the radius $r$ of the circle is $\sqrt{2}$.

Next, we find $OC$, the distance from the center $O(1,1)$ to the midpoint $C$ of $AB$. Since $A(0,2)$ and $B(2,2)$, the midpoint $C$ of $AB$ is:
\[
C=\left(\frac{0+2}{2},\frac{2+2}{2}\right)=(1,2).
\]

Calculate $OC$:
\[
OC=\sqrt{(1-1)^2+(2-1)^2}=\sqrt{0+1}=1.
\]

Next, we calculate the length $AB$:
\[
AB=\sqrt{(2-0)^2+(2-2)^2}=\sqrt{2^2+0^2}=\sqrt{4}=2.
\]

Since $OA=OB=\sqrt{2}$ and $AB=2$, triangle $AOB$ is a right triangle at $O$, so
\[
\angle BOA=90^\circ.
\]

Therefore, the length $OC=1$, and $AB=2$.

So, the correct option is:
\[
\boxed{C}.
\]

\texttt{</think>}
\par\medskip
\noindent $\boxed{C}$ \hspace{5pt}$\leftarrow$ \textcolor{mygreen}{\texttt{Correct}}

\end{minipage}
\end{casebox}
\caption{Qualitative example of PRCO-7B on diagram-grounded geometry reasoning. The Observer localizes points, coordinates, and segment relations from the diagram, and the Solver then uses this structured evidence to derive $\angle BOA$, $OC$, and $AB$.}
\label{fig:case_geometry_reasoning}
\end{figure*}

We present four PRCO-7B cases in Figs.~\ref{fig:case_object_filter}--\ref{fig:case_geometry_reasoning}, covering object filtering, chart reading, table-based selection, and geometry. Each shows the input, the Observer caption, and the Solver response, allowing the intermediate evidence to be compared with the subsequent operations.

In the object-filtering case, the caption preserves attribute combinations rather than object categories alone. The Solver removes one large gray matte cube and two small metallic cylinders from eight objects, while retaining the small brown matte cylinder. Material is therefore needed to distinguish objects with the same shape. The chart case depends on a strict threshold applied to each item: the smallest reported sales value is one, so no item has a store with sales below one. A qualitative description such as low sales would not preserve this decision boundary.

The table case makes question relevance explicit. LLaVA-13B (LLaMA-2) has the highest \texttt{Rec} score, $39.2$, whereas MM-ReAct-GPT-4 leads the \texttt{Total} column. The Observer extracts \texttt{Rec} values together with their model labels; the Solver identifies the maximum and maps the corresponding variant to option F. Preserving model identities matters because the table contains several LLaVA variants with different scores. The required comparison is left to the Solver rather than supplied as a final answer in the caption.

In the geometry case, the caption records $A(0,2)$, $B(2,2)$, $O(1,1)$, and the relation between $OC$ and the midpoint of $AB$. The Solver uses these constraints to obtain $C=(1,2)$, $OC=1$, $AB=2$, and $\angle BOA=90^\circ$. This example separates visually specified coordinates and segment relations from the midpoint, distance, and angle calculations performed downstream.

\begin{figure*}[h]
\centering
\begin{promptbox}{Prompt for Observer}
\textbf{You are the Observer.}

You are given an image and a relevant question. Your task is to write a caption that extracts
only the visually grounded details most useful for answering the question later.

\textbf{Question:} \texttt{\{question\}}

\textbf{Caption guidelines:}
\begin{itemize}[leftmargin=*, label=-, nosep]
  \item Focus on the question’s target(s): describe relevant objects/regions/values with attributes and spatial relations (left/right/top/bottom, near/far, inside/overlapping), including counts and comparisons when relevant.
  \item Include all visually grounded details necessary and sufficient to answer the question later. You may omit visual clutter unrelated to the question.
  \item Write the caption with enough visual evidence that a later LLM can answer Question as if it had seen the image (using only the caption + the question).
  \item Don’t provide the final answer. Only describe what is directly observable in the image, and avoid any additional reasoning or calculations.
  \item Write a thorough caption that preserves enough visual details to reconstruct the scene later; avoid overly short summaries.
  \item When the answer depends on specific labels, numbers, or option mappings, transcribe them explicitly rather than summarizing.
\end{itemize}

Please format the caption as a Markdown bullet list rather than a paragraph.
\end{promptbox}
\begin{promptbox}{Prompt for Solver}
\textbf{You are the Solver.}

Use the caption as the PRIMARY and DEFAULT input and solve directly from it.

Only check the image when the caption does not provide enough information to complete the task.

\textbf{The detailed caption of the provided image:} \texttt{<caption>}

\textbf{Question:} \texttt{\{question\}}

Now perform your reasoning inside \texttt{<think>...</think>}, then output the final answer in \texttt{\textbackslash boxed\{\}}.
\end{promptbox}

\begin{promptbox}{Prompt for GRPO and DAPO}
\{Question\}\\

You first think through the reasoning process as an internal monologue, enclosed within \texttt{<think> </think>} tags. Then, provide your final answer enclosed within \textbackslash boxed\{\}.
\end{promptbox}
\caption{Training and inference prompt templates of the PRCO Observer and Solver, GRPO, and DAPO.}
\label{fig:observer_prompt}
\end{figure*}

\begin{figure*}[h]
\centering

\begin{promptbox}{Prompt for Leakage Checker}
\textbf{You are the Leakage Checker.}

You will be given:
- The question.

- The Observer caption describing an image.

\textbf{Your task:} decide whether the caption contains answer leakage.

\textbf{Question:} \texttt{\{question\}}

\textbf{Observer caption:} \texttt{<caption>}

\textbf{Reminder:} 1 = LEAK, 0 = SAFE.

\textbf{[Leakage Evaluation Rules...]}

\textbf{Output format (MUST follow exactly):}\\
Return a valid JSON object with exactly two keys:
\begin{itemize}[leftmargin=*, label=-, nosep]
  \item \texttt{"label"}: 0 or 1
  \item \texttt{"reason"}: a short reason (1--5 sentences)
\end{itemize}
\textbf{Example:}
\begin{verbatim}
{
  "reason": "The caption does not explicitly state a final answer.",
  "label": 0
}
\end{verbatim}

Now output the final answer as the JSON object (and nothing else).
\end{promptbox}

\begin{promptbox}{Prompt for Error Type Classification}
\textbf{You will be given the following inputs:}
\begin{itemize}[leftmargin=*, label=-, nosep]
  \item \textbf{Image}
  \item \textbf{Question:} \texttt{\{question\}}
  \item \textbf{Model response:} \texttt{\{model\_response\}}
  \item \textbf{Gold answer:} \texttt{\{gold\_answer\}}
\end{itemize}

\textbf{Your task:} Decide which error type best describes why the model response differs from the gold answer, and output exactly one primary label plus a 1--3 sentence rationale.

\textbf{Choose exactly one PRIMARY type from this set:}
\begin{itemize}[leftmargin=*, label=-, nosep]
  \item \textbf{Perception}: Perception error (misread options/tables, failed to extract info; if downstream mistakes are caused by misreading/missing visual info, label as Perception even if reasoning also fails).
  \item \textbf{Reasoning}: Reasoning error (missing/contradictory steps, wrong logical chain, arithmetic/calculation mistakes, algebraic manipulation errors, etc.).
  \item \textbf{Knowledge}: Factual / domain knowledge error.
  \item \textbf{Extraction}: Final-answer extraction/formatting error.
  \item \textbf{Other}: Other / cannot categorize.
\end{itemize}

\textbf{Output requirements:}
\begin{itemize}[leftmargin=*, label=-, nosep]
  \item Return a single minified JSON object only.
  \item Do not output markdown, code fences, or any extra explanation.
  \item Use exactly this format: \texttt{\{"rationale":"1-3 sentences why","category":"<label>"\}}
\end{itemize}
\end{promptbox}
\caption{Prompt templates for leakage checking and error-type classification.}
\label{fig:error_type_classification_prompt}
\end{figure*}

\end{document}